\documentclass[letterpaper, 10 pt, conference]{ieeeconf}
\IEEEoverridecommandlockouts
\usepackage[T1]{fontenc}
\usepackage[utf8]{inputenc} 
\usepackage{microtype}        
\usepackage{amsmath, amssymb}
\usepackage{mathtools}        
\usepackage{graphicx}
\graphicspath{{img/}}         
\usepackage{booktabs}
\usepackage{tabularx}
\newcolumntype{Y}{>{\centering\arraybackslash}X}
\usepackage{multirow}
\usepackage{pifont}

\let\labelindent\relax
\usepackage{enumitem}

\usepackage{siunitx}          
\usepackage[section]{placeins} 
\usepackage{stfloats}        
\usepackage{xurl}        
\usepackage{balance}     
\usepackage{cite}

\usepackage{booktabs}
\usepackage{tabularx}
\usepackage{float}
\usepackage[most]{tcolorbox}
\usepackage{listings}
\tcbset{listing engine=listings}
\usepackage[ruled,vlined]{algorithm2e}
\newcommand{\citet}[1]{\cite{#1}}
\newcommand{\citep}[1]{\cite{#1}}

\title{\LARGE \bf
PCASim: Promptable Closed-loop Adversarial Simulation for Urban Traffic Environment
}

\author{Chuancheng Zhang$^{1, 2 \dag}$, Zhenhao Wang$^{1 \dag}$, Kaizheng Li$^{1,2}$, Yaran Lin$^{1,2}$, Qiang Guo$^{3*}$, Bin Jiang$^{1,2*}$  
\thanks{\dag Both authors contributed equally to this research.}
\thanks{$^{1}$Chuancheng Zhang, Zhenhao Wang, Kaizheng Li, Yaran Lin and Bin Jiang are with Shenzhen Research Institute of Shandong University, Shenzhen, 518057, Guangdong, China}
\thanks{$^{2}$School of Airspace Science and Engineering, Shandong University, 264209 Weihai, China.}
\thanks{$^{3}$Qiang Guo is with the School of Computer Science and 
Technology, Shandong University of Finance and Economics, 250014, Jinan, China.}
\thanks{*Corresponding author email: jiangbin@sdu.edu.cn}
}

\begin{document}
\maketitle
\thispagestyle{empty}
\pagestyle{empty}

\begin{abstract}
    Real-world autonomous driving, particularly in urban environments with numerous corner cases, requires rigorous testing to ensure product safety and robustness. However, few studies have explored integrating adversarial scenario generation with the training of safety agents in closed-loop testing, enabling efficient co-evolution and mutual enhancement of both. To address this challenge, an adversarial behavior knowledge repository is constructed by applying rule-based filtering to an open-source dataset, combined with knowledge retrieval modules tailored for simulation environments. A large language model (LLM) is employed to integrate knowledge-, data-, and adversarial-driven approaches, generating safety-critical traffic scenarios customized to user needs.  Additionally, while evaluating the generated scenarios, we employ reinforcement learning models to train the behaviors of different types of vehicles, thereby enriching scenario diversity beyond existing datasets while preserving realism. Experimental results demonstrate that the proposed framework improves the accuracy of domain-specific language generation by 12\%. Moreover, the success rate of newly generated scenario transformations increases by 8\%, while obstacle-avoidance capability is enhanced by 30\%. For the complete manuscript, please refer to: https://zhenhaooo.github.io/PCASim.github.io/
    
\end{abstract}




\section{Introduction}
Despite significant advancements in autonomous driving technology, high-level autonomous driving accidents, exemplified by Waymo~\citep{WaymoIncidents2024} and Tesla~\citep{TeslaAutopilotFatalities2024}, continue to occur frequently. This underscores the persistent challenges faced in achieving a fully reliable and safe system. In contrast to traditional vehicle safety assessments, which typically rely on controlled collision experiments and behavioral tests within predictable environments, autonomous driving systems (ADS) are designed to operate in far more open and uncontrolled settings. Therefore, prior to large-scale deployment, it is essential to consider a wide range of testing scenarios to include the inherent range of challenges~\citep{gao2022performance}.

The immense diversity of real-world traffic scenarios indicates that relying solely on physical testing is insufficient to address all potential risks~\citep{wei2024risk}. To mitigate this issue, the traditional paradigm involves collecting traffic data from real-world scenes through onboard sensors, which naturally reflect the true distribution of driving data~\citep{wang2024survey}. This data is then used to construct training datasets for developing the driving capabilities of ADS. However, this approach is inefficient and contains many redundant scenarios, which, if used directly for training, could hinder the model's ability to learn the safety-critical scenarios that are more likely to cause higher collision rates~\citep{ding2023survey}. Consequently, increasing attention has been paid to generating adversarial scenarios, such as more hazardous scenarios of lane change~\citep{zhang2025generative} or car follow-up situations~\citep{yin2024kinetic}. Furthermore, current research has demonstrated that training autonomous vehicles with generated adversarial scenarios can significantly reduce safety issues in ADS~\citep{wang2021advsim, feng2023trafficgen}.

\begin{figure*}[ht]
  \centering
  \includegraphics[width=1.0\textwidth, trim=0 80 0 40, clip]{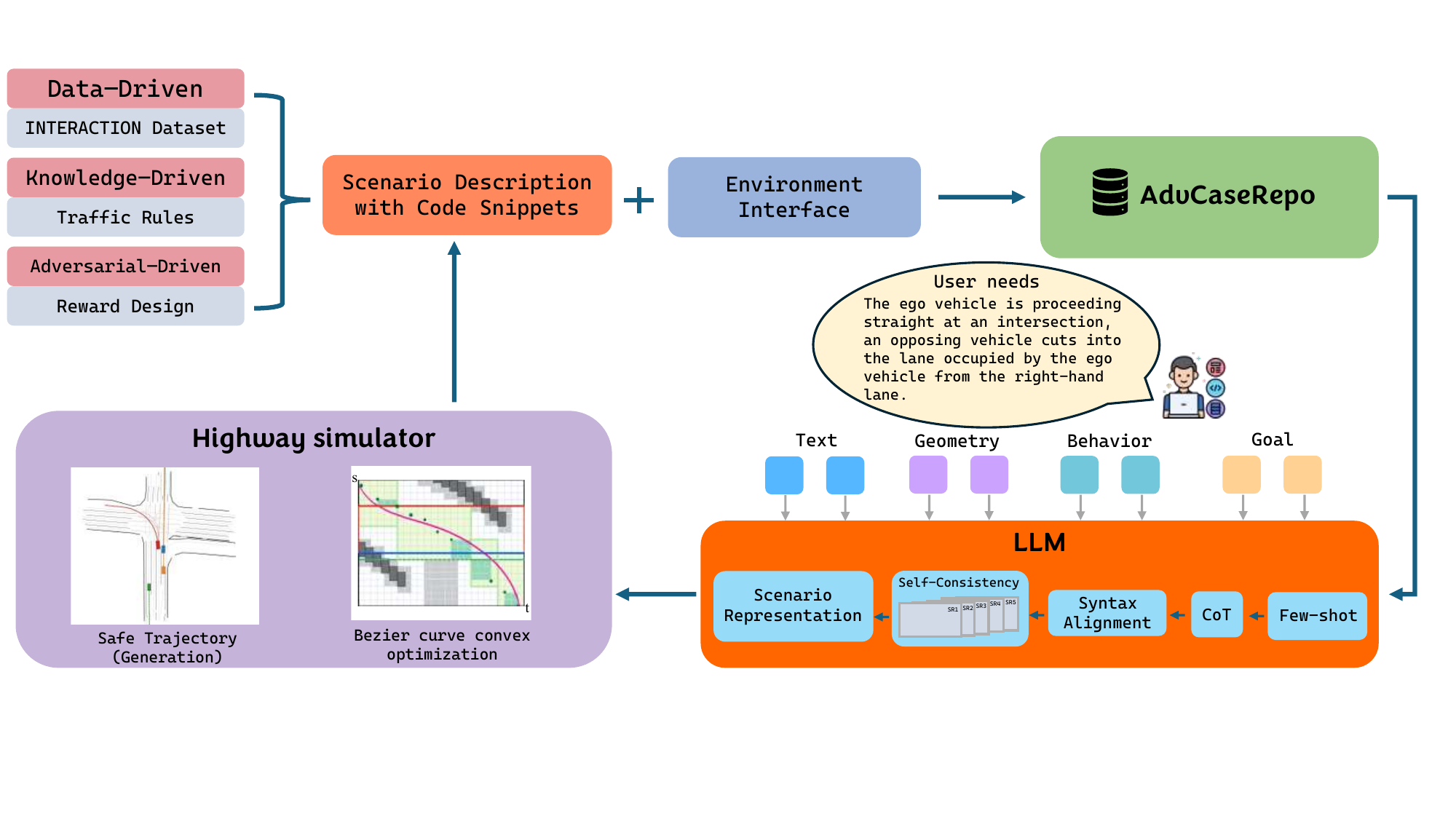}
  \caption{The overall framework of the closed-loop adversarial simulation 
  }
  \label{fig:example}
  \vspace{-0.1in}
\end{figure*}

Currently, safety-critical scenario generation methods, represented by data-, knowledge- and adversarial-driven approaches — excel in different stages such as scene understanding, rule creation and encoding. These methods often fail to integrate effectively and fully capitalize on their respective strengths. However, with the advent of LLMs trained on massive internet-scale data and billions of parameters, remarkable capabilities in commonsense reasoning, planning, interaction and decision-making have been demonstrated~\citep{touvron2023llama}, highlighting their immense potential to address the aforementioned challenges. This paper proposes a novel closed-loop traffic simulation framework, aimed at bridging the gap between data-driven and knowledge-driven approaches, with a focus on the design of the data end and the enhancement of LLM's adversarial scenario generation capabilities. By requiring only natural language descriptions from the user, it utilizes specialized designs for LLM to generate a Domain-Specific Language (DSL) that guides the step-by-step execution of each dynamic object within the scene, thereby creating complex adversarial scenarios for the robust and safe training of autonomous vehicles. To fully utilize these scenarios, a reinforcement learning agent is trained to avoid adversarial behaviors embedded within the environment, resulting in complete scenarios encompassing both adversarial actions and successful avoidance maneuvers. 

As shown in Fig. 1, we have developed a rule-based refinement method to extract potential hazardous scenarios from an open-source urban intersection dataset~\citep{zhan2019interaction}, which further contributes to the construction of a scenario corpus. This corpus consists of primary scenes centered on the ego vehicle and secondary scenes centered on the adversarial vehicle. Based on this corpus and leveraging current mainstream large language models~\citep{guo2025deepseek, liu2024deepseek, yang2024qwen2}, we employ Retrieval-Augmented Generation (RAG) to assist in retrieval, further enhanced by prompt engineering and reinforcement learning-based optimization to construct a comprehensive adversarial scenario repository. Moreover, a lightweight middleware translates the generated DSL into Python code, enabling integration with the simulation environment. To the best of our knowledge, this represents the first notable attempt in closed-loop ADS testing to achieve on-demand scenario generation and leverage the extended scenarios for training autonomous agents. In summary, the contribution of this paper is three-fold:


\begin{enumerate}

\item An urban traffic adversarial scenario repository was constructed, and a RAG+LLM-based prompt engineering paradigm was developed for scenario generation. This framework enables efficient, diverse, and interpretable generation of realistic and strongly adversarial scenarios tailored to specified prompts.

\item An RL-based traffic-flow model is proposed to control the ego and adversarial vehicles, simultaneously validating the quality of the generated adversarial scenarios and producing new safety-critical scenario data.

\item A closed-loop scenario generation and evaluation framework is introduced to assess and filter high-quality scenarios, iteratively enriching the scenario repository and alleviating the limitations of small-sample distributions.
\end{enumerate}    
    

\section{Related Works}

\subsection{Generation of Safety-Critical Scenarios}
Safety-critical scenario generation for autonomous vehicles (AVs) typically falls into three categories: data-, adversarial-, and knowledge-based approaches. Data-driven approaches~\citep{Tan2023,Scanlon2021,Yang2020} advocates the utilization of real-world data to guide vehicle behavior. \citet{Wei2024}  proposed extracting static and dynamic variables from the CIMSS-TA dataset and employing conditional tabular generative adversarial networks (CTGANs) to increase diversity of generated scenarios. Although realistic, this approach is limited by long-tail sparsity of critical scenarios and costly data collection.
Traditional adversarial-driven approaches often utilize Monte Carlo Tree Search~\citep{Lee2015}, Bayesian optimization~\citep{Abeysirigoonawardena2019}, and reinforcement learning within the framework of structured domain randomization (SDR)~\citep{Prakash2019}. Grounded in uncertainty-aware, sampling-based policy optimization, these techniques balance exploration and exploitation in sparse-reward settings and generate realistic adversarial scenarios.
However, these methods are often inefficient and generate adversarial environments with limited diversity. To address these limitations, A generative adversarial network (GAN)-based model combined with reinforcement learning, incorporating human driving priors to generate naturalistic adversarial scenarios, was proposed by \citep{Hao2023}. \citet{hanselmann2022king} introduced kinematics gradients to generate robust, safety-critical scenarios for imitation learning.  \citet{Mei2025} introduced a closed-loop adversarial scenario generation framework based on LLM. This framework iteratively refines scenarios, ultimately producing high-quality adversarial environments with significantly enhanced generation efficiency. \citet{chen2024frea} proposed feasibility-guided generation, balancing adversariality and realism in safety-critical scenarios. Knowledge-based generation produces scenarios based on predefined traffic rules and physical constraints~\citep{Bagschik2018,Cai2020,Klischat2020}. Previous studies rely on complex rule encoding, yet manually crafted rules could not cover all safety-critical cases.
With the advancement of end-to-end approaches in autonomous driving, \citet{Deng2023} introduced TARGET — an end-to-end architecture that automatically generates ADS test scenarios based on traffic rules, effectively addressing the complexity of rule-to-simulator mapping and minimizing manual intervention.

\subsection{LLM for Autonomous Driving Simulation}
LLM exhibits human-like understanding and reasoning capabilities in complex scenarios. For instance, \citet{Deng2023} validated LLMs’ capabilities in knowledge extraction, verification, and syntactic alignment, enabling them to comprehend contextual information and generate structured outputs, effectively simulating human reasoning in complex scene interpretation. \citet{Tian2024} introduced a large multimodal model (LMM) capable of reasoning, abstraction and scenario construction from traffic videos using a multimodal few-shot chain-of-thought (CoT) approach. Through leveraging these capabilities, LLMs are increasingly used for generating and optimizing safety-critical scenarios. \citet{Sun2024} further improved ADS training by incorporating reinforcement learning with human feedback into the LLM-based framework. ChatScene utilizes LLM to automatically extract data, generate and retrieve code, render and evaluate scenes and iteratively optimize collision-related parameters, thereby producing safety-critical scenarios for ADS development~\citep{Zhang2024}. Further, Text2Scenario presented in \citep{Cai2025} is the most closely related to our method, which employs LLMs to autonomously generate simulation test scenario frameworks from user-provided natural language descriptions. This approach enables the creation of segmented scenarios within simulation environments and significantly enhances the efficiency of autonomous driving testing. Moreover, the closed-loop ProSim framework together with its self-constructed driving dataset exhibits strong performance~\citep{tan2024promptable}, but it lacks a systematic quality evaluation of the generated scenarios.

\section{Methodology}

To overcome the limitations of isolated scenario generation methods, we propose a unified framework
for scenario construction and trajectory optimization in closed-loop simulation under LLM guidance. As shown in Figure~\ref{fig:example}, the system transforms descriptions into structured DSL via a RAG process, compiles them into simulation code and refines ego trajectories with convex optimization. To enhance scenario diversity, adversarial agents are trained in the generated scenes. After training, the adversarial agent’s weights are frozen, and the ego vehicle is subsequently trained against these fixed behaviors. Successful avoidance cases are appended to the corpus, creating an iterative training loop.

\subsection{Scenario Corpus Construction via Multi-Driven Fusion}

Our corpus construction begins with the INTERACTION dataset~\citep{zhan2019interaction}, containing detailed vehicle trajectories at urban intersections. Rule-based extraction methods identify first-level ego vehicle scenarios, such as driving straight, turning, braking, following, and lane changes. Considering interactions between the ego vehicle and background vehicles, adversarial scenario categories are further refined based on background vehicle behaviors, e.g., a straight-moving ego vehicle encountering a left-turning background vehicle. Scene classification uses spatiotemporal features like relative heading, time-to-collision (TTC), post-encroachment time (PET), and lane index transitions. Symbolic traffic knowledge from road network data adds lane properties such as line types, driving directions, and temporary control signals. These features are translated into descriptions (e.g., “temporary road closure,” “slow-speed lane with dashed markings”) that serve as a knowledge-driven scaffold for understanding the environment.
In parallel, adversarial conditions are synthesized by adding behaviors like sudden braking, tailgating, unsafe lane changes, and excessive speeding. Each adversarial vehicle is positioned relative to the ego vehicle with behavior parameters reflecting challenging conditions, e.g., tailgating at 0.5m distance from the rear, or lateral cut-in at 1.2m from the left. 
This generation expands the corpus' behavioral diversity, enhancing its relevance for testing robustness.

The outputs from these three sources in Fig.\ref{fig:example} are unified into a structured intermediate representation. Each scenario consists of five elements: a scene type label, a data-driven behavioral summary, a knowledge-driven road description, an adversarial condition descriptor, and a natural language scene description. This representation forms the basis for prompt-driven generation, allowing the LLM to condition its output on multiple semantic dimensions. The final corpus is serialized as an Excel file, with trajectory snippets and metadata preserved for further parsing. This corpus serves as the input for the retrieval-augmented DSL generation process and forms the primary semantic layer of the adversarial scenario repository.

\subsection{DSL Generation with RAG-Augmented LLM}

Generating executable DSL from natural descriptions provided by users based on their specific scenario requirements presents a dual challenge: capturing fine-grained semantic context and preserving structural correctness. RAG is adopted to address these challenges. Unlike traditional vector-space models (VSMs) that rely on surface-level similarity, RAG leverages dense embeddings and example-grounded generative reasoning, thereby improving retrieval fidelity and inference accuracy. This capability is particularly valuable in setting, where scenario fidelity depends not only on matching relevant scene elements, but also on composing structurally valid and contextually appropriate DSL code for downstream execution.

A DSL retrieval corpus is built from the processed scenarios. Each entry contains a structured scene description and its corresponding DSL representation, divided into geometry, generation, and behavior. They are mapped to the road network, positions and adversarial behaviors. The records are embedded using Sentence-BERT~\citep{reimers2019sentence} and indexed via FAISS~\citep{johnson2019billion}. During inference, a new description is transformed into an embedding to retrieve the top-$k$ relevant entries.
The generation process employs a CoT prompt, facilitating step-by-step reasoning to generate well-structured DSL. The prompt template incorporates few-shot examples and a scenario component dictionary to guide vocabulary selection and syntax usage. DeepSeek-V3 is employed as the base model due to its reasoning and interactive performance.
A semantic alignment verification step is introduced to ensure the semantic fidelity of the generated DSL. The generated elements are first verified against the key attributes and behaviors described in the input natural language scene. If inconsistencies are detected, the model revises the output accordingly. Subsequently, the generated DSL elements are cross-checked against the predefined scenario components in the adversarial scenario repository. If no close match is found, the output is retained as long as it remains consistent with the original scene description. This alignment process guarantees both semantic accuracy and syntactic validity before code generation.
To enhance robustness, a self-consistency mechanism~\citep{wang2022self} samples the LLM multiple times per query and uses embedding-based voting to select the most consistent DSL, reducing syntax drift. The finalized DSL, enriched with retrieved scenario elements and code snippets, is input to a code-generation module that produces Python compatible with modified \texttt{highway-env}.

\begin{figure*}[t]
  \centering
  \includegraphics[width=0.9\textwidth, trim=-50 0 0 0, clip]{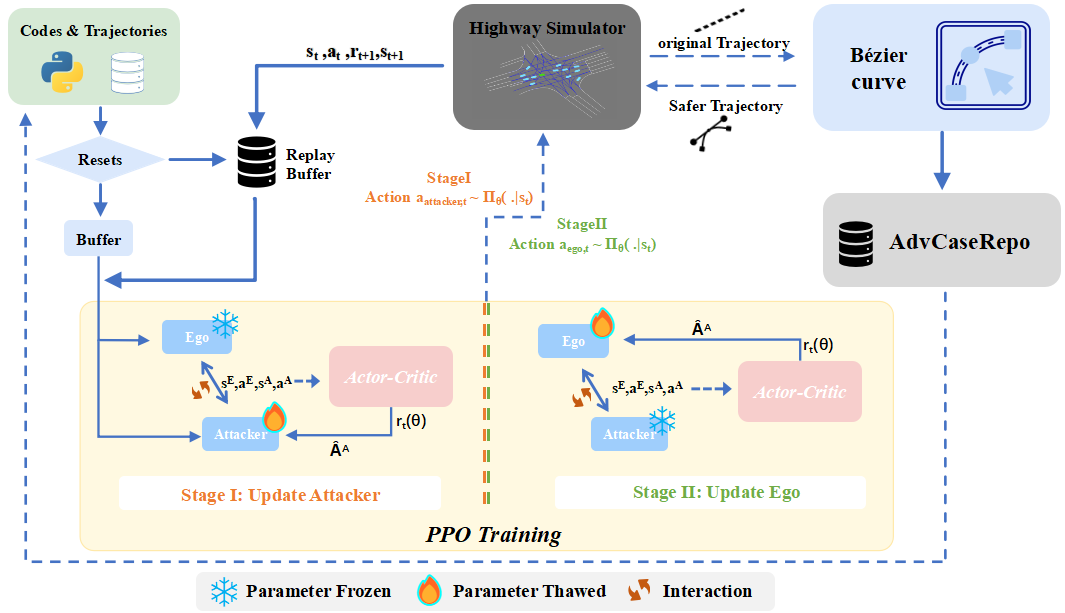}
  \caption{The framework of newly generated adversarial scenarios.}
  \label{fig:ppo}
  \vspace{-0.1in}
\end{figure*}

\subsection{Adversarial Scenario Repository and Code Generation}

To bridge the gap between DSL generation and executable code, an adversarial scenario repository is constructed. It functions as a capability layer consisting of a semantic retrieval base and a code generation interface. Unlike static corpora, the repository is designed as a dynamic and executable resource collection that integrates domain-specific scenario templates, high-level behavior taxonomies, and vectorized code anchors, thereby supporting multi-stage reasoning and simulation integration. The repository consists of three major parts: (i) a scenario corpus containing natural language descriptions paired with manually or semi-automatically generated DSL representations, segmented into geometry, spawn and behavior modules; (ii) a semantic dictionary and scenario taxonomy that define hierarchical component types, enabling controlled generation and alignment verification; and (iii) a FAISS-indexed database of Python code fragments extracted from the modified \texttt{highway-env} environment. Each entry in the repository is embedded using Sentence-BERT and annotated with metadata, including scenario intent, vehicle roles, and interaction patterns. 

During DSL generation, the repository supports few-shot prompting and component grounding. Given a new scenario description, the top-$k$ semantically similar scene pairs are retrieved to guide CoT reasoning. The hierarchical dictionary constrains vocabulary usage, ensuring syntactic correctness and semantic alignment.  

By combining scene abstraction with code-level representation, the repository provides a reusable, extendable, and executable knowledge base that supports high-quality scenario generation and simulation. This design ensures semantic consistency, structural completeness, and enables scalable integration of new scenario types and simulator features. The repository is periodically augmented with successful adversarial-training cases (see Section~\ref{3.4}).

\subsection{Adversarial Agent Training and Corpus Augmentation}
\label{3.4}

To improve the diversity, difficulty and realism of traffic scenarios, we establish a closed-loop simulation cycle that alternates between adversarial training, ego vehicle adaptation, trajectory refinement, and corpus augmentation (as shown in Fig.~\ref{fig:ppo}). This cycle refines agent behaviors and incrementally expands the scenario repository with new data from simulation feedback, enhancing scenario diversity beyond existing datasets while maintaining realism. 

\paragraph{Adversarial and Ego Agent Training.} 
Based on adversarial vehicle selections from the original trajectory segments, a PPO-based reinforcement learning model is employed to train adversarial behaviors, as shown in Fig~\ref{fig:ppo}. An adversarial reward function maintains vehicle aggressiveness and a reset technique ~\citep{nikishin2022primacy} reinitializes planner parameters to prevent convergence to local optima. After adversarial agent training is completed, the adversarial vehicle model is fixed and the ego vehicle is trained using a specifically designed reward function to ensure safe decision-making. 
\paragraph{Trajectory optimization and scenario recollection.} Bézier curve convex optimization is applied to smooth trajectories, reducing excessive lateral acceleration while preserving the original heading constraints.
Collision avoidance is prioritized, followed by trajectory planning and execution of the strategy. Successful interactions between ego and adversarial vehicles are retained as closed-loop simulation outputs, thus expanding the repository’s coverage of adversarial intentions and response patterns. 

Our realism evaluation aligns with the metrics used in the Waymo Sim Agent Challenge, emphasizing kinematic feasibility via Bézier smoothing and interaction fidelity through distribution matching of TTC and PET metrics. This process establishes a self-improving simulation system, where agent robustness and scenario repository are enhanced through two coupled feedback loops: a behavioral loop strengthening robustness and a data loop enriching the corpus with new high-quality examples.

\begin{figure}[!t]
  \centering
  \includegraphics[width=1\linewidth]{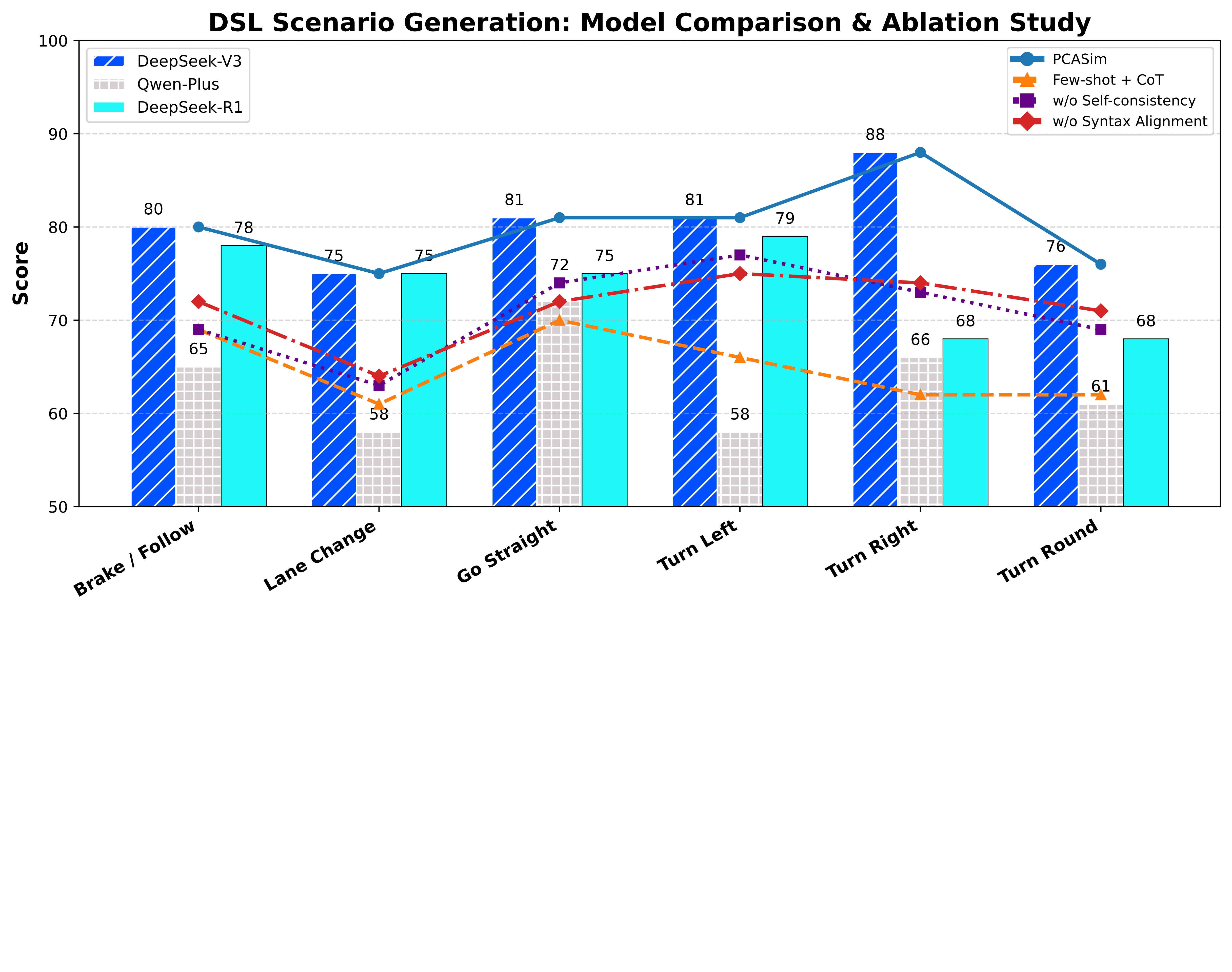}
  \vspace{-1.25in}
  \caption{\textbf{Comparison} of DSL scenario generation scores across different LLMs and \textbf{ablation} study of core components in the DSL generation pipeline.}
  \label{fig:dsl_score}
  \vspace{-0.2in}
\end{figure}

\section{Experiment}


\subsection{Experiment Setup}



Our experimental framework evaluates the fidelity of generated DSL scenes and assesses the safety-critical performance of the autonomous agent through three key tasks. Firstly, conduct cross-model comparisons among different LLMs to identify their reliability in generating accurate and executable DSL scenarios. An ablation study further investigates the impact of prompt engineering and voting mechanisms, highlighting their contributions to scenario accuracy. Next, generated DSL scenarios are compiled into executable Python scripts and evaluated using a PPO-trained agent within our customized highway-env, focusing on behaviors such as going straight, braking, lane changing, and turning. Performance metrics, including collision rates, timeout occurrences, and conversion success rates, are systematically logged and analyzed across multiple runs to ensure robustness. 
\subsection{Experimental Results}


\subsubsection{Comparison of LLM Models under DSL Generation Metrics}
We first evaluate how different LLMs perform in generating structured DSL scenes from natural language descriptions. For this, we apply the DSL scoring criteria introduced in Section~4.5, which includes semantic fidelity, executable validity, structural completeness, modularity, behavioral richness and voting centrality. Each LLM (e.g., DeepSeek-V3, Qwen2.5-Plus, DeepSeek-R1) is prompted with identical scenario descriptions using our full pipeline (few-shot + CoT + alignment + voting) and the average scores are reported over a held-out test set of 6 diverse scenes.

As shown in Fig.~\ref{fig:dsl_score}, DeepSeek-V3 consistently achieves the highest DSL generation scores across all ego behavior categories, with an average score of 80.17, outperforming both DeepSeek-R1 (73.83) and Qwen2.5-Plus (63.33). Its advantage is particularly pronounced in more complex behaviors such as Turn Right and Turn Left, where reasoning about interaction logic and scene structure is critical. This suggests that DeepSeek-V3 is better equipped to handle fine-grained semantic composition and structural alignment under our retrieval-augmented and prompt-guided framework.
Notably, while DeepSeek-R1 performs comparably in scenarios like Brake/Following and Lane Change, its performance fluctuates more across behaviors, reflecting less stability under diverse traffic configurations. Qwen2.5-Plus, on the other hand, consistently lags behind, indicating limitations in maintaining semantic and syntactic coherence despite being exposed to the same few-shot CoT prompt format.

These results highlight that the choice of LLM substantially influences the quality and fidelity of the driving scenarios generated. This underscores the importance of LLM selection for scenario generation tasks in closed-loop autonomous driving simulations.

\subsubsection{Comparison with baseline}
\begin{figure}[htb]
  \centering
  \vspace{-0.5em}
  \includegraphics[
    width=\linewidth,
    trim={20 100 60 120},clip
  ]{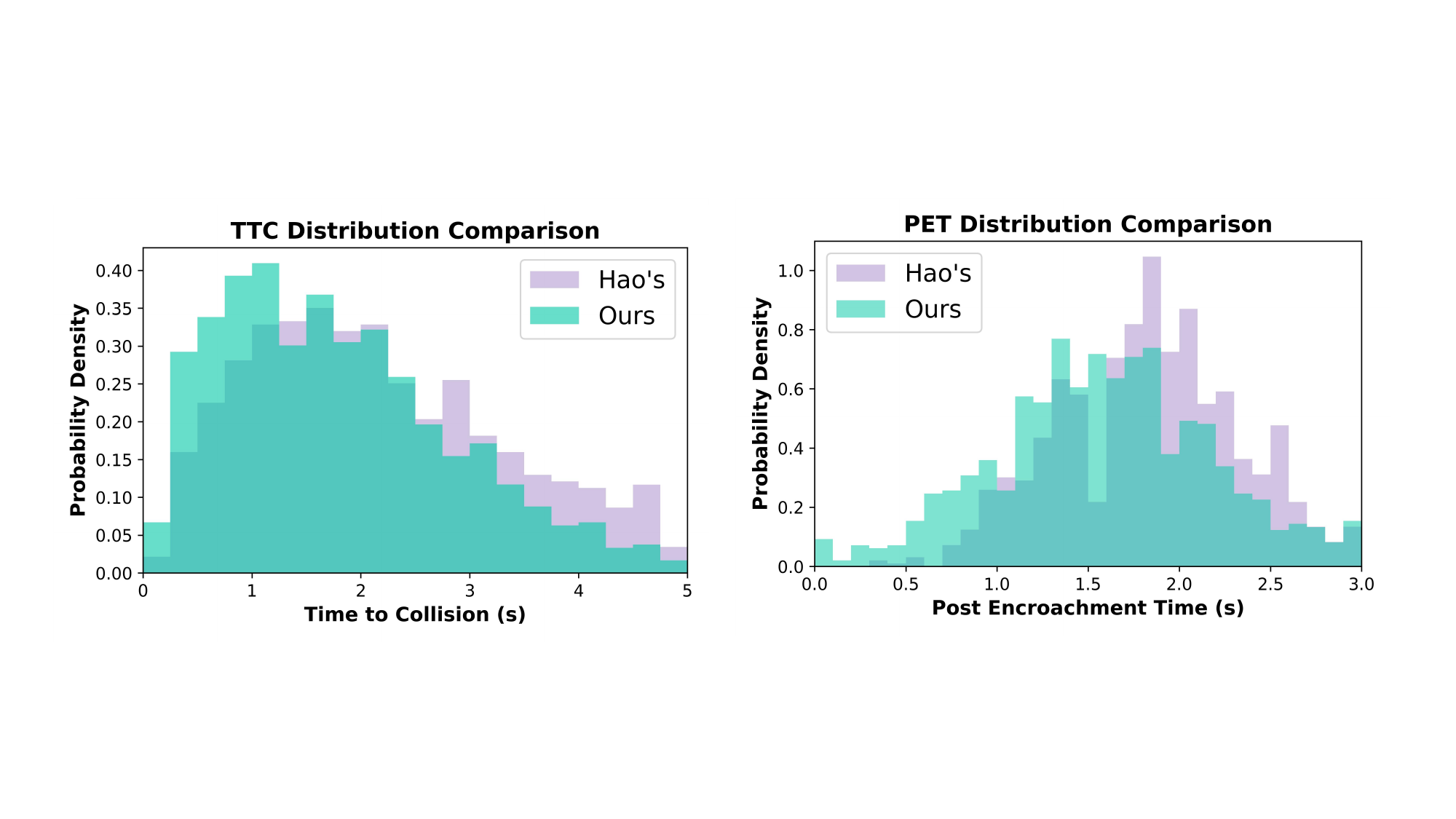}
  \vspace{-2.5em}
  \caption{Evaluation under adversarial criteria}
  \label{fig:1}
\end{figure}

\begin{figure}[htb]
  \centering
  \vspace{-0.5em}
  \includegraphics[
    width=0.8\linewidth,
    trim={10 10 30 10},clip
  ]{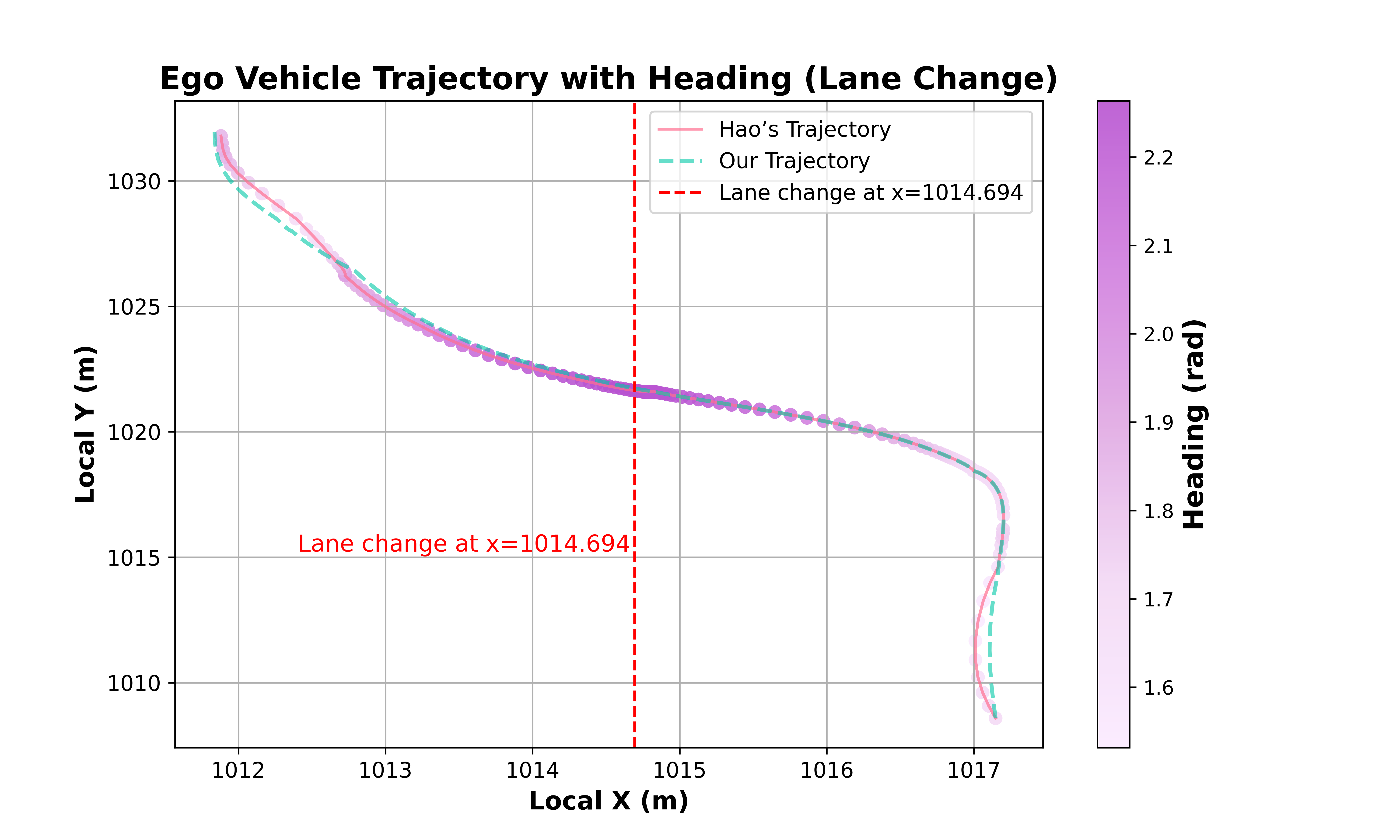}
  \vspace{-1.0em}
  \caption{Comparison of the greatest heading trajectories}
  \vspace{-1em}
  \label{fig:2}
\end{figure}

Since the target is not limited to a single platform and the datasets differ, experiments were not conducted on cases such as KING~\citep{hanselmann2022king} and FREA~\citep{chen2024frea}. Instead, based on the same simulator and dataset, we selected the method described in Hao’s work~\citep{Hao2023} as the benchmark. Comparisons of generated scenarios in both Fig.~\ref{fig:1} and Table~\ref{tab:1} with respect to adversarial metrics such as TTC, PET, and collision rate (CR) clearly demonstrate that our method achieves superior distributions. Notably, PCASim achieves an increase of about 8\% in collision rate over baseline method, indicating its enhanced effectiveness in generating challenging adversarial scenarios (Phase I). Moreover, as shown in Fig.~\ref{fig:2}, the maximum heading angles of RL-validated trajectories enhanced by Bézier smoothing are significantly improved, indicating that our approach produces more natural trajectories and achieves enhanced sim-to-real transferability in scenario resampling (Phase II).

\begin{table}[htb]
  \centering
\caption{Collision rate comparison}
\label{tab:1}
  \begin{tabular*}{\linewidth}{@{\extracolsep{\fill}} lcc}
    \toprule
    Method & Phase~I$\uparrow$ & Phase~II$\downarrow$ \\ \midrule
    Baseline & 62.84\% & 42.17\% \\
    Ours     & \textbf{70.78\%} & \textbf{39.93\%} \\ \bottomrule
  \end{tabular*}
\end{table}

\subsubsection{Ablation Study on Prompting and Alignment Mechanisms}
To quantify the contribution of semantic alignment and self-consistency, we perform ablation experiments using DeepSeek-V3. Each configuration's performance is measured by the DSL score across a standardized set of scenarios.

\begin{table*}[t]
  \caption{Ablation study of core components in the DSL generation pipeline.}
  \label{tab:ablation}
  \centering
  \setlength{\tabcolsep}{6pt}        
  \renewcommand{\arraystretch}{1.15} 
  \fontsize{7pt}{10pt}\selectfont

  \begin{tabularx}{\textwidth}{lYYYYYYY}
    \toprule
    \textbf{Configuration} & \textbf{Brake/Follow} & \textbf{Lane Change} & \textbf{Go Straight} & \textbf{Turn Left} & \textbf{Turn Right} & \textbf{Turn Round} & \textbf{Average} \\
    \midrule
    \textbf{PCASim}                & 80 & 75 & 81 & 81 & 88 & 76 & \textbf{80.17} \\
    \textbf{Few-shot + CoT}        & 69 & 61 & 70 & 66 & 62 & 62 & \textbf{64.83} \\
    \textbf{w/o Self-consistency}  & 69 & 63 & 74 & 77 & 73 & 69 & \textbf{70.83} \\
    \textbf{w/o Syntax Alignment}  & 72 & 64 & 72 & 75 & 74 & 71 & \textbf{71.33} \\
    \bottomrule
  \end{tabularx}

  \vspace{2pt}
  \small{
  \textit{*} “w/o" stands for “without", indicating the removal of the respective component.
}
\end{table*}

\setlength{\textfloatsep}{8pt plus 2pt minus 2pt}
\begin{figure*}[t]
  \centering
  \includegraphics[width=1.0\textwidth, trim=35 100 45 80, clip]{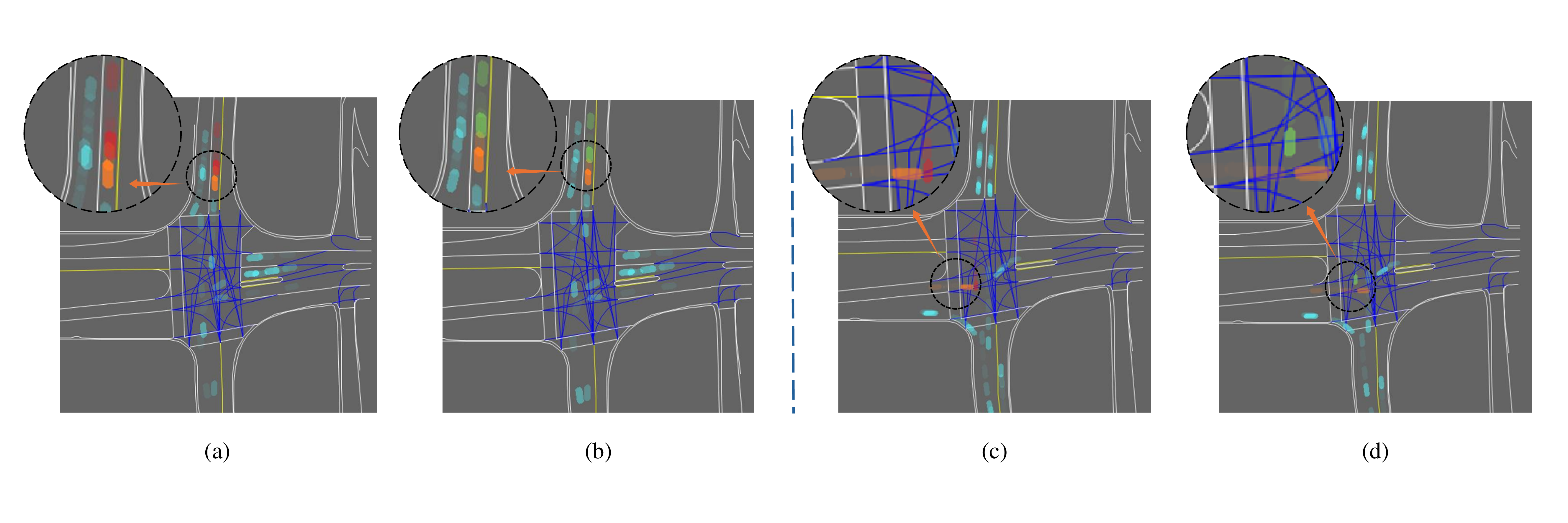}
  \caption{\textbf{Generating Scenarios and Validation.} (a) (b) are the braking scenarios, (c) (d) are the going straight scenarios, the orange indicates adversarial vehicle, the red indicates ego vehicle without training and the green ego vehicle has been strengthened by the reinforcement learning.}
  \label{fig:view}
\end{figure*}

Fig.~\ref{fig:dsl_score} presents an ablation study that evaluates the contribution of each core component in our DSL generation pipeline. It is clear to observe that the complete pipeline, which includes few-shot prompting, CoT reasoning, semantic alignment and self-consistency voting, achieves the highest overall performance with an average DSL score of 80.17, showing a 12\% improvement in the accuracy of domain-specific language generation compared to configurations without semantic alignment, as shown in Table~\ref{tab:ablation}. This indicates that while few-shot prompting and reasoning are foundational for structured generation, they are insufficient alone to ensure high semantic fidelity and executable validity.
The moderate score drop (70.83) following the removal of self-consistency suggests its primary contribution lies in stability. In contrast, the absence of semantic alignment (71.33) reveals its critical impact on DSL-to-description fidelity, especially for intricate behaviors like left and U-turns.

These results underscore a modular synergy: few-shot prompting and CoT establish a robust generative foundation, while alignment and voting act as critical post-hoc safeguards to rectify inconsistencies and suppress variance. Altogether, the full pipeline ensures both structural integrity and semantic correctness in the generated scenarios, confirming the necessity of integrating reasoning, validation and aggregation in closed-loop DSL generation. Specifically, the RAG and semantic alignment modules ensure the 'direction' by grounding DSL to user prompts, while the Phase I RL training injects the 'adversarial' nature by optimizing conflict intensity.

\subsubsection{Collision and Timeout Metrics in Adversarial Scenarios}
To evaluate adversarial strategy learning in multi-vehicle scenarios, we designed a training and evaluation framework that tracks collision and timeout rates in four scenario types: braking, going straight, turning, and lane changing. Multiple scenes are sampled during each training session and run in parallel using a shared PPO model. Trained vehicles are initialized in their own environment and controlled by the PPO policy, while other vehicles follow a fixed behavior. After collecting trajectories for all scenarios, we aggregate the data and perform a policy update. This centralized update strategy ensures consistent learning while benefiting from multi-process rollout acceleration. This framework enables systematic large-scale adversarial training and provides metrics (mean performance) for evaluating the impact of adversarial behavior under various realistic traffic conditions. As shown in Fig.~\ref{fig:view}, by introducing PPO combined with a reset technique for training, the ego vehicle demonstrates safer obstacle avoidance capabilities against adversarial vehicles compared to the untrained baseline. 

\begin{table}[th]
\caption{Collision rate and timeout statistics under adversarial settings.}
\label{tab:collision-stats}
\centering
\begingroup
\setlength{\tabcolsep}{4.5pt}   
\renewcommand{\arraystretch}{1.05} 
\footnotesize                    
\begin{tabular}{@{}lcc@{\hspace{6pt}}cc@{}} 
\toprule
\multirow{2}{*}{\textbf{Ego Behavior}} &
\multicolumn{2}{c}{\textbf{Adv against ego}} &
\multicolumn{2}{c}{\textbf{Ego avoid adv}} \\
\cmidrule(lr){2-3}\cmidrule(r){4-5}  
 & \textbf{Collision Rate} & \textbf{Timeout}
 & \textbf{Collision Rate} & \textbf{Timeout} \\
\midrule
Brake       & 72.7\% & 3 & 46.1\% & 7 \\
Lane Change & 57.1\% & 3 & 24.2\% & 6 \\
Go Straight & 73.3\% & 4 & 46.6\% & 8 \\
Turn        & 80.0\% & 2 & 42.8\% & 8 \\
\bottomrule
\end{tabular}
\par\medskip
\scriptsize \emph{Note.} Each experimental metric is taken from the average of multiple runs.
\endgroup
\end{table}

The trained ego vehicle exhibits markedly lower collision rates than the untrained baseline — demonstrating greater robustness and adaptability in adversarial scenarios of lane change, straight-driving, and turning. In Table~\ref{tab:collision-stats}, the avoidance policy lowers collision rates by roughly 30 \% on average while maintaining stable timeout counts, thus boosting safety without sacrificing efficiency.

\subsubsection{Comparative Studies with Recent Related Work}
\label{subsubsection:comparative_study}
\newcommand{\cmark}{\ding{51}}  
\newcommand{\xmark}{\ding{55}}  

To provide context for our framework, we conducted a comparative analysis with recent related works in autonomous-driving scenario generation, as summarized in Table \ref{tab:scenario_comparison}. 

\begin{table}[htbp]
\centering
\caption{Comparison of scenario-generation frameworks for autonomous-driving simulation.}
\label{tab:scenario_comparison}
\fontsize{6pt}{10pt}\selectfont
\begin{tabular}{lcccc}
\toprule
\textbf{Method} &
\textbf{Promptable} &
\textbf{Validation} &
\textbf{Training Aug.} &
\textbf{Scene Repo (self).} \\
\midrule
Text2Scenario \cite{Cai2025} & \cmark & \xmark & \xmark & \cmark \\
ProSim \cite{tan2024promptable} & \cmark & \xmark & \xmark & \cmark \\
CAT \cite{zhang2023cat} & \xmark & \xmark & \cmark & \xmark \\
VCAT \cite{xuancai2024vcat} & \xmark & \xmark & \cmark & \xmark \\
\textbf{Ours} & \cmark & \cmark & \cmark & \cmark \\
\bottomrule
\end{tabular}

\end{table}

While frameworks like Text2Scenario and ProSim enable prompting for scene creation, they operate without a closed-loop validation mechanism, which limits their reliability for safety-critical testing. Text2Scenario converts natural language to OpenScenario DSL in an open-loop manner, and ProSim focuses on multi-modal realism rather than stress testing. Conversely, CAT and VCAT are designed to generate adversarial scenarios to improve system robustness. For instance, CAT accelerates adversarial synthesis, while VCAT achieves a high attack success rate by using vulnerability-aware rewards. Our framework distinguishes itself by integrating all these capabilities: it combines natural-language prompting with closed-loop validation and automated scene exploration without needing a pre-built repository. This unique combination allows our system to explore both nominal and extreme traffic conditions, yielding a 12\% gain in DSL accuracy and a 30\% boost in executable scene yield. This integrated approach ensures both controllability and validation, making our framework effective for comprehensive testing.

\section{Limitations}

Our framework is subject to several important limitations that constrain its generalizability and scalability. The scenario corpus is currently constructed from the INTERACTION dataset~\citep{zhan2019interaction}, which, despite its richness in urban interactions, provides limited geographical and behavioral diversity. The simulation environment relies on \texttt{highway-env}, which has a simplified kinematic model facilitates efficient experimentation but lacks high-fidelity dynamics and multimodal sensing capabilities. Furthermore, the pipeline depends on externally hosted LLM APIs (e.g., GPT-5~\citep{openai_gpt4o_2024}, Claude 4~\citep{anthropic_claude_2024}), which introduces uncertainties in availability, stability, and scalability due to external constraints.

Methodological limitations include DSL conversion errors due to map format mismatches and API-driven inconsistencies in LLM generation. Moreover, insufficient traffic-rule reasoning and the RL's narrow focus on basic maneuvers—omitting complex multi-stage dynamics—collectively diminish the diversity and robustness of the resulting system.

Future work will focus on mitigating these limitations by incorporating heterogeneous datasets (e.g., NGSIM, nuScenes, Waymo) to broaden scenario diversity, migrating to high-fidelity simulators such as CARLA~\citep{Dosovitskiy17}, and developing domain-specific LLMs with enhanced traffic reasoning.


\section{Conclusion}

This paper presents PCASim, a promptable closed-loop simulation framework for generating and evaluating safety-critical urban traffic scenarios. The framework incorporates an adversarial scenario repository together with a RAG+LLM-based prompt engineering paradigm, enabling efficient, diverse, and interpretable scenario generation tailored to specified prompts. An RL-based traffic-flow model is employed to jointly control the ego and adversarial vehicles, which not only validates the fidelity of the generated scenarios but also produces additional safety-critical data. Moreover, a closed-loop generation–evaluation pipeline is designed to iteratively filter and enrich the repository, alleviating the limitations of small-sample distributions and ensuring comprehensive coverage of challenging traffic events.

Compared with existing approaches, PCASim achieves a higher degree of directedness and controllability in adversarial scenario generation. By leveraging prompt-guided DSL translation with semantic alignment, the framework generates qualitatively more adversarial and diverse cases than rule-based or purely generative methods. In combination with PPO-based reinforcement learning and Bézier-curve convex optimization, the proposed system substantially improves the robustness of autonomous vehicles, equipping them with stronger capabilities to navigate complex and adversarial urban environments.

\section{Acknowledgement}
The study is funded by the Shenzhen Fundamental Research Program (JCYJ20230807094104009).




\bibliographystyle{IEEEtran}
\bibliography{example} 

\newpage
\appendix

\section{Corpus to Python}

\subsection{Initialize Corpus and Generate DSL}
\begin{figure*}[t] 
  \centering
  \includegraphics[angle=90, height=1.0\textwidth, width=0.95\textheight, keepaspectratio, trim=20 150 0 40, clip]{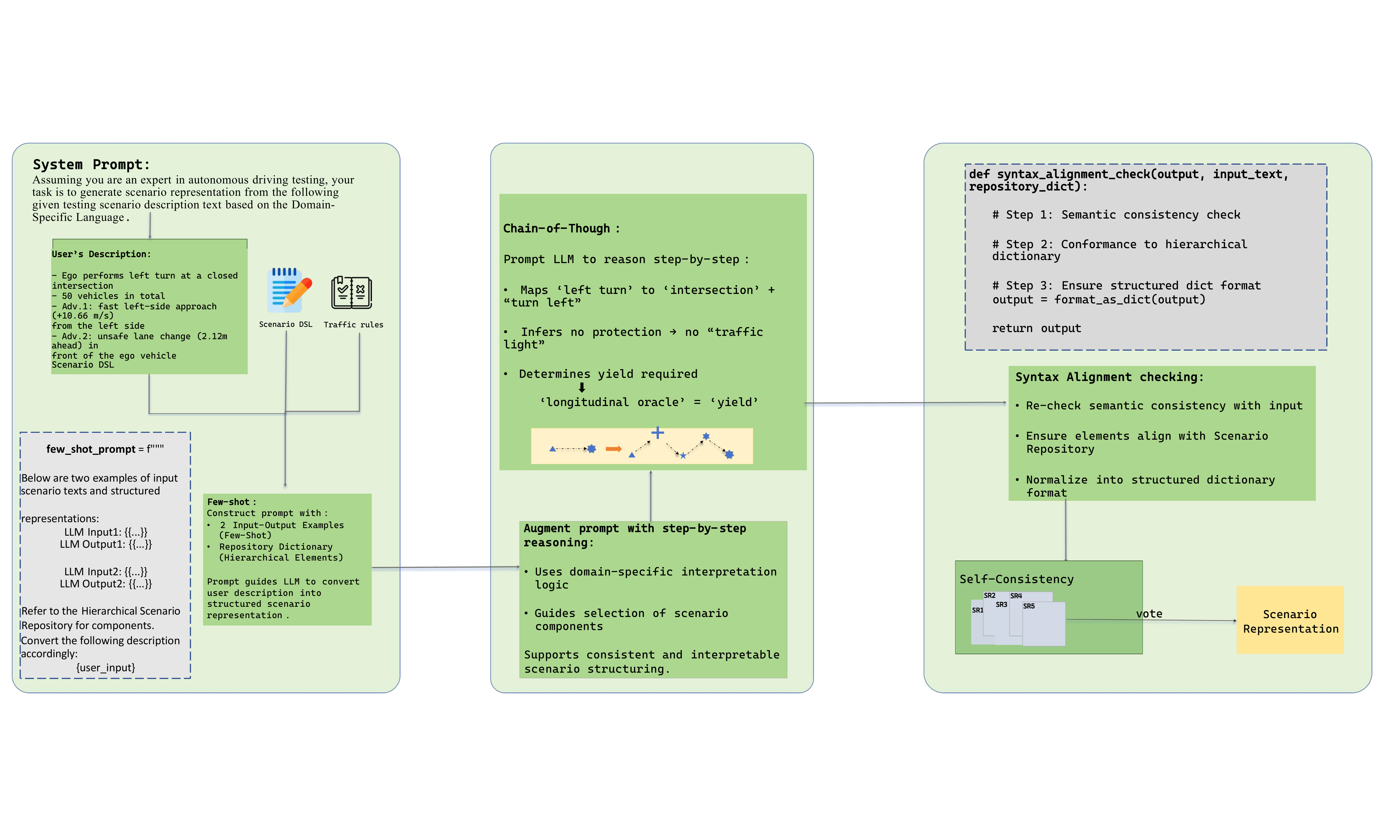}
  \caption{Framework of retrieval-augmented prompt construction, Chain-of-Thought reasoning, syntax alignment checking, and self-consistency voting for DSL generation.}
  \label{fig:example2}
\end{figure*}

To better illustrate the overall logical flow from prompt construction to scenario representation generation, the corresponding framework is visualized in Figure~\ref{fig:example2}. The process is then detailed in the following subsections.

\subsubsection{How to Deal with the Initial Open-source Dataset - Interaction}

The initial data source utilized in this work is the \textbf{INTERACTION} dataset, which provides real-world trajectories of vehicles and pedestrians at urban intersections. To effectively extract structured driving scenarios from this raw data, a comprehensive preprocessing and feature extraction pipeline is applied, comprising the following steps:

\paragraph{Data Preprocessing.}
The raw CSV files containing vehicle and pedestrian trajectories are first cleaned to handle missing values and outliers. The data is then transformed into a structured dictionary format to facilitate subsequent processing. Each vehicle and pedestrian trajectory is segmented based on timestamps and the resulting fragments are saved as JSON files for efficient access and further analysis.

\paragraph{Feature Extraction.}
Key motion features---including velocity, acceleration and yaw rate---are extracted to characterize driving behaviors. Using OpenStreetMap (OSM) road network data, each vehicle's positional information is matched to corresponding lane identifiers. This enables the classification of different driving maneuvers based on spatial and temporal features:

\begin{itemize}
    \item \textbf{Car-following behavior:} Identified by detecting the closest preceding vehicle within the ego-vehicle's trajectory frames, ensuring that at least one vehicle is in motion. Metrics such as following duration and time-to-collision (TTC) are computed.
    \item \textbf{Braking behavior:} Recognized within car-following segments by detecting continuous deceleration events lasting at least five frames.
    \item \textbf{Lane change behavior:} Determined by comparing consecutive lane indices; a lane change is confirmed when the current lane index differs from the previous frame and both belong to adjacent lanes in the map.
    \item \textbf{Intersection maneuvers (straight, left turn, right turn, U-turn):} Identified by analyzing trajectory intersections between the ego-vehicle and background vehicles, computing post-encroachment time (PET) and total yaw angle changes to classify the type of maneuver.
\end{itemize}

\paragraph{Scenario Classification.}
Based on the extracted behaviors, trajectories are clustered into high-level and fine-grained scenario categories. High-level scenarios include car-following, braking, lane changing and intersection maneuvers. Fine-grained subcategories further distinguish combinations such as ``ego vehicle going straight with background vehicle turning left,'' ``left-turn car-following,'' and ``right-turn braking.''

\paragraph{Risk Assessment.}
Each extracted scenario is quantitatively assessed using TTC and PET metrics to evaluate its potential risk level, enriching the diversity of the dataset and enhancing its applicability for downstream learning tasks.

\paragraph{Scenario Data Output.}
The final output of this extraction process is a structured collection of scenario data stored in \texttt{JSON} format. Each record contains detailed trajectory information, interaction labels, risk metrics and scenario categories. These JSON files serve as the foundational corpus for integrating knowledge-driven, data-driven and adversarial-driven insights into a unified natural language description framework, which subsequently enables downstream DSL generation.

\subsubsection{Integrate knowledge-, data- and adversarial-driven information into a complete natural description}

To bridge the gap between raw structured data and downstream DSL generation, we integrate information derived from data-driven, knowledge-driven and adversarial-driven sources into a unified natural language description. The integration process is designed as follows:

\paragraph{Data-driven Insight Extraction.}
The ego vehicle's motion patterns are summarized by parsing the processed JSON trajectory data. Behaviors such as braking, following, lane changes and intersection crossings are extracted and translated into concise natural language summaries, providing a realistic foundation based on observed driving behaviors.

\paragraph{Knowledge-driven Translation.}
Environmental context is inferred from the underlying OpenStreetMap (OSM) road network. Information about lane types, temporary road changes (e.g., closures or detours) and general traffic facilities is synthesized into complementary descriptions. This enriches the scenario understanding beyond vehicle-centric observations.

\paragraph{Adversarial-driven Extension.}
Adversarial elements are simulated and injected into the scene descriptions. These include stochastic behaviors such as sudden braking, unsafe lane changes, speeding and tailgating, along with their relative spatial relationships to the ego vehicle. The generated adversarial conditions introduce critical, safety-relevant perturbations to the otherwise naturalistic scenarios.

\begin{algorithm}[th]
\caption{Assemble Descriptions from Scenario Files}
\label{alg:assemble_descriptions}
\KwIn{Folder of JSON files $F$, Road Network File $R$, Scenario Type $S$}
\KwOut{Excel File with Final Descriptions}

\SetKwFunction{FExtract}{ExtractTrajectorySnippets}
\SetKwFunction{FAssemble}{AssembleDescription}
\SetKwFunction{FProcess}{ProcessFolder}

\BlankLine
\textbf{Function} \FExtract{$J$}:\\
\Indp
    Load JSON file $J$\\
    Parse ego vehicle's trajectory info\\
    Extract: behavior, vehicle IDs, start/end time\\
    \Return trajectory snippets list\\
\Indm

\BlankLine
\textbf{Function} \FAssemble{$S, J, R, D, N$}:\\
\Indp
    $desc.\text{scenario\_type} \leftarrow S$ \\
    $desc.\text{data\_driven} \leftarrow$ GenerateDataInsight($J$)\\
    $desc.\text{knowledge\_driven} \leftarrow$ GenerateKnowledgeTranslation($R$)\\
    $adv \leftarrow$ GenerateAdversarialExtension(return\_json=True)\\
    $desc.\text{adv\_description} \leftarrow adv.\text{natural\_description}$\\
    $desc.\text{adv\_vehicles} \leftarrow$ JSONEncode($adv.\text{adversarial\_vehicles}$)\\
    $desc.\text{final\_natural\_language} \leftarrow$ Combine all descriptions + dataset info\\
    $desc.\text{trajectory\_snippets} \leftarrow$ \FExtract{$J$} \\
    \Return $desc$\\
\Indm

\BlankLine
\textbf{Function} \FProcess{$F, R, S, O$}:\\
\Indp
    Initialize $results \leftarrow [~]$\\
    \ForEach{file $f$ in folder $F$}{
        \If{$f$ ends with .json}{
            $J \leftarrow$ full path of $f$ \\
            $desc \leftarrow$ \FAssemble{$S, J, R, F, f$}\\
            Append $desc$ to $results$
        }
    }
    Save $results$ to Excel file $O$\\
\Indm

\BlankLine
\textbf{Main}:\\
\Indp
    Set paths for folder $F$, map $R$, and output $O$\\
    \FProcess{$F, R, S, O$}
\end{algorithm}

\paragraph{Unified Natural Language Assembly.}
The above three sources are programmatically fused into a final descriptive paragraph. A typical assembled description includes:
\begin{itemize}
    \item The ego vehicle's extracted behavior pattern.
    \item Environmental insights derived from map and infrastructure knowledge.
    \item Adversarial vehicle behaviors and their relative positions.
    \item Dataset provenance information (e.g., original file names and data sources).
\end{itemize}
This complete natural language description serves as the direct input for downstream DSL generation modules, ensuring that both the factual realism and adversarial diversity of driving scenarios are preserved.The detailed pseudocode for the integration process is provided in Algorithm \ref{alg:assemble_descriptions}

\subsubsection{Construct the corpus}

After generating structured natural language descriptions, we proceed to construct a structured corpus to support downstream DSL generation. This corpus encapsulates not only polished textual descriptions but also modular DSL representations segmented into geometry, spawn and behavior components. The construction process is organized as follows:

\paragraph{Input Natural Descriptions.}
The process begins with a collection of assembled natural language descriptions, each corresponding to a specific driving scenario. These descriptions integrate data-driven, knowledge-driven and adversarial-driven elements, serving as the foundation for further modularization.

\paragraph{Refinement and Modularization.}
Each initial description undergoes a refinement process guided by a language model, aiming to enhance linguistic fluency and simulation-oriented clarity while strictly preserving semantic fidelity. Simultaneously, the descriptions are decomposed into three modular snippets:
\begin{itemize}
    \item \textbf{geometry.snippet}: Defines the ego vehicle's initial location and intended movement direction, assuming a generalized road network structure.
    \item \textbf{spawn.snippet}: Specifies the initial setup of all vehicles involved, including types, relative positions and orientations.
    \item \textbf{behavior.snippet}: Describes the sequential behaviors of the ego vehicle and surrounding traffic participants over the scenario timeline.
\end{itemize}

\paragraph{Few-shot Prompt Augmentation.}
To ensure generation consistency and quality, a dynamic memory bank of few-shot examples is maintained. Selected high-quality snippet examples are injected into the prompting process, guiding the language model to produce structurally and semantically aligned outputs. The memory bank is continuously updated as the corpus expands.

\paragraph{Corpus Storage.}
The final corpus is organized into an Excel file, \texttt{final\_snippet\_export.xlsx}, with the following sheets:
\begin{itemize}[nosep, leftmargin=*]
    \item \texttt{description}: The refined complete scenario descriptions.
    \item \texttt{geometry.snippet}: Geometry-level DSL snippets.
    \item \texttt{spawn.snippet}: Spawn-level DSL snippets.
    \item \texttt{behavior.snippet}: Behavior-level DSL snippets.
\end{itemize}

\paragraph{Prompt Design.}
The prompt structure employed to guide the generation process emphasizes strict semantic alignment, modular output requirements and adherence to predefined simulation environment standards. An illustrative example of the designed prompt and corresponding model response is provided in Table~\ref{tab:corpus_prompt_response}.

\begin{table*}[t]
\centering
\caption{Example of corpus construction prompt and model response.}
\label{tab:corpus_prompt_response}
\begin{tcolorbox}[colback=white!5!white, colframe=black!75!black, title=Prompt:]
\textbf{System Prompt:}  
You are an autonomous driving simulation expert. Please complete the following task.

\textbf{Input Prompt:}
\begin{enumerate}
    \item Polish the Natural Language Scenario Description:  
    Transform the following raw natural language driving scenario description into a clear, natural and simulation-oriented English narrative in paragraph form. Semantic consistency must be strictly maintained—no changes to the original intent are allowed.
    
    \item Structure Requirements:
    \begin{itemize}
        \item \textbf{geometry.snippet} should define the first-level scene, including the ego vehicle's location and movement direction. It is not necessary to specify the full road network structure; instead, provide a template that references reading from an \texttt{.osm} map. Additionally, clearly specify the source dataset on which this scenario is based and annotate the dataset name in brackets, e.g., [INTERACTION].
        \item \textbf{spawn.snippet} should define the initial vehicle setup, including the type, relative positions and orientations of the ego vehicle and any adversarial vehicles.
        \item \textbf{behavior.snippet} should describe the behaviors of both the ego vehicle and adversarial vehicles throughout the scenario.
    \end{itemize}

    \item Ensure that the semantic content of all three DSL modules matches the polished natural language description. Each element in the DSL must have a clear one-to-one semantic correspondence to the narrative description.

    \item Each section must begin with import or reference statements based on the existing base classes of our project \texttt{highway-env}. Avoid introducing excessive custom classes to ensure structural consistency and maintainability. For example:
\begin{verbatim}
from NGSIM_env import utils
from NGSIM_env.envs.common.abstract import AbstractEnv
from NGSIM_env.road.road import Road, RoadNetwork
from NGSIM_env.vehicle.behavior import IDMVehicle
\end{verbatim}

You can refer to the following code to understand the predefined classes and their usage: \{rag\_snippets\}.

The following is a previously generated example of a DSL module (used as a memory repository): \{few\_shot\_examples\}.

\textbf{Please output in the following format:}

[Polished Description]

[geometry.snippet] (DSL block)

[spawn.snippet] (DSL block)

[behavior.snippet] (DSL block)
\begin{itemize}
    \item Original Description
    \item Based on real dataset
\end{itemize}
\end{enumerate}
\end{tcolorbox}

\vspace{0.5em}

\begin{tcolorbox}[colback=white!5!white, colframe=black!75!black, title=Response:]
\begin{verbatim}
sheets = {
    "description": pd.DataFrame({"description"}),
    "geometry.snippet": pd.DataFrame({"geometry.snippet"}),
    "spawn.snippet": pd.DataFrame({"spawn.snippet"}),
    "behavior.snippet": pd.DataFrame({"behavior.snippet"}),
}
\end{verbatim}
\end{tcolorbox}
\end{table*} %

The detailed pseudocode for the corpus construction process is provided in Algorithm \ref{generation of corpus}.

\begin{algorithm}[th]
\caption{DSL Scenario Generation via Prompting and Retrieval}
\label{generation of corpus}
\KwIn{Scene description file $D$, highway-env code base $C$, initial snippet library $T$}
\KwOut{Excel file with generated snippets, updated template JSON}

\SetKwFunction{FExtract}{ExtractSnippet}
\SetKwFunction{FPrompt}{GeneratePrompt}
\SetKwFunction{FLoadIndex}{LoadOrBuildCodeIndex}
\SetKwFunction{FFormat}{FormatFewShotExamples}
\SetKwFunction{FQuery}{QueryLLM}
\SetKwFunction{FSave}{SaveOutputs}

\BlankLine
Load API key and paths\\
Load template library $T$ (geometry/spawn/behavior) from JSON\\
Load DSL corpus $D$ from Excel file\\
Initialize output lists for DSL and refined text\\
$index \leftarrow$ \FLoadIndex{$C$} \tcp*[f]{Load or build FAISS index from code base}

\BlankLine
\ForEach{row $(o, f)$ in corpus $D$}{
    $F_{shot} \leftarrow$ \FFormat{$T$} \tcp*[f]{Construct few-shot examples}\\
    $RAG \leftarrow$ Top-$k$ similarity search over $index$ using $o$\\
    $prompt \leftarrow$ \FPrompt{$o$, $F_{shot}$, $RAG$} \\
    
    \tcp{Query the language model}
    $response \leftarrow$ \FQuery{$prompt$} \\
    
    Extract: $desc$, $geo$, $spawn$, $behav$ using \FExtract{} \\
    
    Append results to output lists\\
    Update $T$ with new snippets if not in template
}

\BlankLine
\FSave{results to JSON, updated $T$ to JSON, Excel multi-sheet file}
\end{algorithm}

\subsubsection{DSL Scenario Generation According To User's Need}

To enable flexible scenario generation tailored to diverse user needs, we design a Retrieval-Augmented Generation (RAG) pipeline combined with self-consistency voting mechanisms. Given an arbitrary natural language scenario description as input, the system retrieves relevant corpus entries, formats a context-enhanced prompt, queries a language model multiple times and consolidates the outputs to generate a consistent and executable DSL representation. The overall pipeline is organized as follows:

\paragraph{Main Pipeline.}
The main generation workflow is outlined in Algorithm~\ref{main_pipeline}. The system first loads the DSL corpus and builds a semantic retriever for context retrieval. Upon receiving a user query, a Chain-of-Thought (CoT) based prompt is constructed. Multiple candidate DSL outputs are sampled via the language model. Depending on the voting mode selected---embedding-based or structure-clustered---the final DSL output is determined by either selecting the most semantically central sample or performing fuzzy-field clustering and voting. The final DSL is then converted into executable Python code for the customized \texttt{highway-env} simulation environment.

\begin{algorithm}[th]
\caption{DSL Generation via RAG and Self-Consistency Voting}
\label{main_pipeline}
\KwIn{Natural language description $q$, DSL corpus $D$, scenario repository $R$}
\KwOut{Executable Python code for highway-env}

\SetKwFunction{FBuild}{BuildRetriever}
\SetKwFunction{FPrompt}{FormatPrompt}
\SetKwFunction{FInvoke}{QueryLLM}
\SetKwFunction{FEmbedVote}{EmbeddingVote}
\SetKwFunction{FStructVote}{StructureClusterVote}
\SetKwFunction{FConvert}{ConvertDSLToCode}

\BlankLine
Load DSL corpus $D$ and build retrieval index via \FBuild{}\\
Define Chain-of-Thought Prompt using repository $R$\\
Construct \texttt{RetrievalQA} chain with "stuff" mode\\

\BlankLine
\textbf{Function} \FEmbedVote{$q$, $chain$, $k$}:\\
\Indp
    Sample $k$ outputs via \FInvoke{$q$}\\
    Compute pairwise cosine similarity between outputs\\
    Select most “central” output as final DSL\\
    \Return final DSL\\
\Indm

\BlankLine
\textbf{Function} \FStructVote{$q$, $chain$, $k$}:\\
\Indp
    Sample $k$ outputs via \FInvoke{$q$}\\
    Parse each DSL into field dictionary\\
    Perform fuzzy-cluster voting to find most frequent field value\\
    Select closest matching DSL as final output\\
    \Return final DSL\\
\Indm

\BlankLine
\textbf{Main}:\\
\Indp
    Choose voting mode (embedding or structure)\\
    \uIf{embedding}{
        $dsl \leftarrow$ \FEmbedVote{$q$, chain, $k$}
    }\Else{
        $dsl \leftarrow$ \FStructVote{$q$, chain, $k$}
    }
    $code \leftarrow$ \FConvert{$dsl$} \\
    Print or save generated $code$
\end{algorithm}

\paragraph{Retriever Construction.}
To efficiently retrieve context-relevant DSL snippets, the corpus is first processed into prompt-style blocks containing integrated descriptions, geometry, spawn, and behavior modules. These text blocks are segmented, embedded using a sentence-transformer model and stored in a FAISS vector index for similarity-based retrieval. The retriever construction process is detailed in Algorithm~\ref{alg:build_retriever}.

\begin{algorithm}[th]
\caption{BuildRetriever: Construct FAISS-based Semantic Retriever}
\label{alg:build_retriever}
\KwIn{Raw DSL scene corpus $D$}
\KwOut{Retriever object for similarity-based lookup}

\SetKwFunction{FSplit}{SplitDocuments}
\SetKwFunction{FEmbed}{EmbedChunks}
\SetKwFunction{FStore}{BuildFAISSStore}

\BlankLine
\textbf{Step 1: Format Corpus into Prompt Blocks}\\
Initialize empty list $B$\\
\ForEach{scene $(d, g, s, b)$ in $D$}{
    $b_i \leftarrow$ Concatenate $d$, $g$, $s$, $b$ into a prompt-style text block\\
    Append $b_i$ to $B$
}

\BlankLine
\textbf{Step 2: Chunk and Embed}\\
$C \leftarrow$ \FSplit{$B$, chunk\_size=800, overlap=100}; \\
$E \leftarrow$ \FEmbed{$C$, model=MiniLM} \tcp*{Use HuggingFaceEmbeddings}

\BlankLine
\textbf{Step 3: Build Vector Store and Retriever}\\
$S \leftarrow$ \FStore{$E$} \tcp*{Build FAISS index}
\Return $S$ as retriever
\end{algorithm}

\paragraph{Prompt Formatting.}
Upon retrieval, the system constructs a structured prompt incorporating the retrieved context, the user query, and Chain-of-Thought reasoning instructions. Syntax alignment guidelines are also included to enforce compliance with the hierarchical scenario repository. Few-shot examples are applied to improve output consistency. The prompt formatting process is detailed in Algorithm~\ref{alg:format_prompt} and the prompt structure for DSL generation is detailed in Table~\ref{tab:dsl_generation_prompt_response}.

\begin{table*}[t]
\centering
\caption{Example of generating DSL representations from natural language descriptions using hierarchical repository guidance and CoT reasoning.}
\label{tab:dsl_generation_prompt_response}
\begin{tcolorbox}[colback=white!5!white, colframe=black!75!black, title=Prompt:]

\textbf{System Prompt:}  
Assuming you are an expert in autonomous driving testing, your task is to generate scenario representations from the following testing scenario description based on the Domain-Specific Language (DSL).

\textbf{Input Prompt:}
\begin{enumerate}
    \item \textbf{Hierarchical Scenario Repository Guidance:}  
    The Hierarchical Scenario Repository provides a dictionary of scenario components corresponding to each subcomponent. When creating scenario representations, please prioritize choosing from existing elements. If no appropriate element is found, you may create a new one.

    \item \textbf{Few-Shot Examples:}
    \begin{itemize}
        \item \textbf{LLM Input 1}: \{\{Input example\}\} \quad \textbf{LLM Output 1}: \{\{Output example\}\}
        \item \textbf{LLM Input 2}: \{\{Input example\}\} \quad \textbf{LLM Output 2}: \{\{Output example\}\}
    \end{itemize}

    \item \textbf{Main Conversion Task:}  
    Based on the above description and examples, convert the following testing scenario text into the corresponding scenario representation: \{\{Input Scenario\}\}

    \item \textbf{Chain of Thought and Syntax Alignment:}
    \begin{itemize}
        \item Think step by step to reason about appropriate mappings from text descriptions to scenario components.
        \item Syntax alignment checking:
        \begin{enumerate}
            \item Ensure semantic consistency between the scenario description and generated scenario representation.
            \item Verify that all elements used are from the Scenario Repository dictionary. If no matching element exists, generate a consistent custom element.
            \item Ensure the output is formatted into a valid dictionary data structure.
        \end{enumerate}
    \end{itemize}

    \item \textbf{Retriever Context (Optional):}  
    \texttt{[Scene i]} \quad \{description\} \quad \texttt{[Scenic Geometry]} \quad \{geometry\} \quad \texttt{[Scenic Spawn]} \quad \{spawn\} \quad \texttt{[Scenic Behavior]} \quad \{behavior\}
\end{enumerate}

\end{tcolorbox}

\vspace{0.5em}

\begin{tcolorbox}[colback=white!5!white, colframe=black!75!black, title=Response:]

\textbf{DSL for User's Need:}
\begin{verbatim}
#------geometry.snippet------#
{
    ... (generated geometry snippet) ...
}

#------spawn.snippet------#
{
    ... (generated spawn snippet) ...
}

#------behavior.snippet------#
{
    ... (generated behavior snippet) ...
}
\end{verbatim}

\end{tcolorbox}
\end{table*} %

\begin{algorithm}[th]
\caption{FormatPrompt: Generate Prompt with CoT and Syntax Check}
\label{alg:format_prompt}
\KwIn{Natural language query $q$, retrieved context blocks $C$, repository dictionary $R$, few-shot examples $F$}
\KwOut{Prompt $P$ for input to LLM}

\BlankLine
Initialize prompt header as expert instruction block\\
Append section: \texttt{CONTEXT BLOCK: \{C\}} \\
Append section: \texttt{LLM Input: \{q\}}\\
Append section: \texttt{LLM Output: \{\{Scenario Representation\}\}}\\

\BlankLine
\textbf{Step 1: Add CoT Reasoning Instructions}\\
Append reasoning hints about road type, yield logic, intersection rules\\

\BlankLine
\textbf{Step 2: Add Syntax Alignment Rules}\\
(1) Check semantic match between description and DSL\\
(2) Check compliance with repository $R$\\
(3) Format output into structured dictionary format\\

\BlankLine
\textbf{Step 3: Insert Few-Shot Examples}\\
Append two complete pairs of LLM Input/Output in expected format from $F$\\

\Return final prompt $P$
\end{algorithm}

\paragraph{Self-Consistency Voting.}
To mitigate stochastic variability in the language model outputs, we implement two self-consistency voting strategies:
\begin{itemize}
    \item \textbf{Embedding-Based Voting:} Multiple DSL outputs are sampled, embedded and compared via pairwise cosine similarity. The output with the highest average similarity to all others is selected as the final DSL.
    \item \textbf{Structure-Clustered Voting:} Each DSL output is parsed into structured fields. A fuzzy clustering algorithm groups similar outputs and the most common field values are voted to reconstruct a consistent DSL representation.
\end{itemize}

\paragraph{Implementation.}
The complete pipeline, including retriever construction, prompt assembly, LLM querying and voting selection, is implemented in the corresponding Python script. It leverages the \texttt{langchain}, \texttt{sentence-transformers} and \texttt{FAISS} libraries to manage retrieval, embedding and voting operations.

\subsection{Initialize Repository and Generate Codes}
\label{appendix:repository}

\subsubsection{Get Repository}

To bridge the gap between DSL generation and executable code, we construct a comprehensive adversarial scenario repository that serves as both a semantic retrieval base and a dynamic code generation backbone. Unlike static corpora, this repository is designed as an extendable and executable resource collection, integrating domain-specific scenario templates, high-level behavior taxonomies, and code-level anchors. Its structure supports multi-stage reasoning, semantic alignment, and simulator integration.

The repository consists of three major components:

\begin{itemize}
    \item \textbf{Scenario Corpus.}  
    A structured corpus containing natural language descriptions paired with manually or semi-automatically generated DSL representations. Each DSL is segmented into three modules: \texttt{geometry.snippet}, \texttt{spawn.snippet}, and \texttt{behavior.snippet}, enabling fine-grained retrieval and controlled scenario assembly.

    \item \textbf{Semantic Dictionary and Scenario Taxonomy.}  
    A hierarchical ontology defining scenario component types, behavior categories, and vehicle interaction patterns. This taxonomy enables constraint-based generation, semantic validation, and alignment checking during both DSL generation and code conversion.

    \item \textbf{Executable Code Repository.}  
    A FAISS-indexed database of Python code fragments extracted from the modified \texttt{highway-env} simulation environment. Each code snippet corresponds to specific DSL elements, such as map layouts, vehicle initialization, and behavior controllers. The snippets are embedded using a Sentence-BERT model to enable semantic similarity retrieval during DSL-to-code translation.
\end{itemize}

Each entry in the repository is annotated with metadata, including scenario intent labels, vehicle roles, and dynamic interaction information.  
The retrieval system leverages semantic similarity metrics to dynamically fetch relevant code examples, which are incorporated into the prompt construction for the language model.

This repository design ensures:

\begin{itemize}
    \item \textbf{Semantic Consistency:} The mapping between DSL representations and executable code maintains semantic fidelity across abstraction levels.
    \item \textbf{Structural Completeness:} Generated Python scripts include all essential modules, ensuring simulation readiness.
    \item \textbf{Extensibility:} New scenario types, behavior classes, and simulation modules can be incrementally added to the repository without retraining or manual relabeling.
\end{itemize}

Technically, the repository construction is automated by parsing the simulation codebase, segmenting files into semantically coherent fragments, embedding them using sentence-transformer models, and organizing them into FAISS vector indices for efficient retrieval. Supplementary tools for nearest-neighbor matching of DSL descriptions are also provided, supporting flexible corpus augmentation and similarity search during evaluation.

\subsubsection{DSL-to-Python Code Generation}

\begin{figure*}[ht]
  \centering
  \includegraphics[width=1.12\textwidth, trim=250 250 0 40, clip]{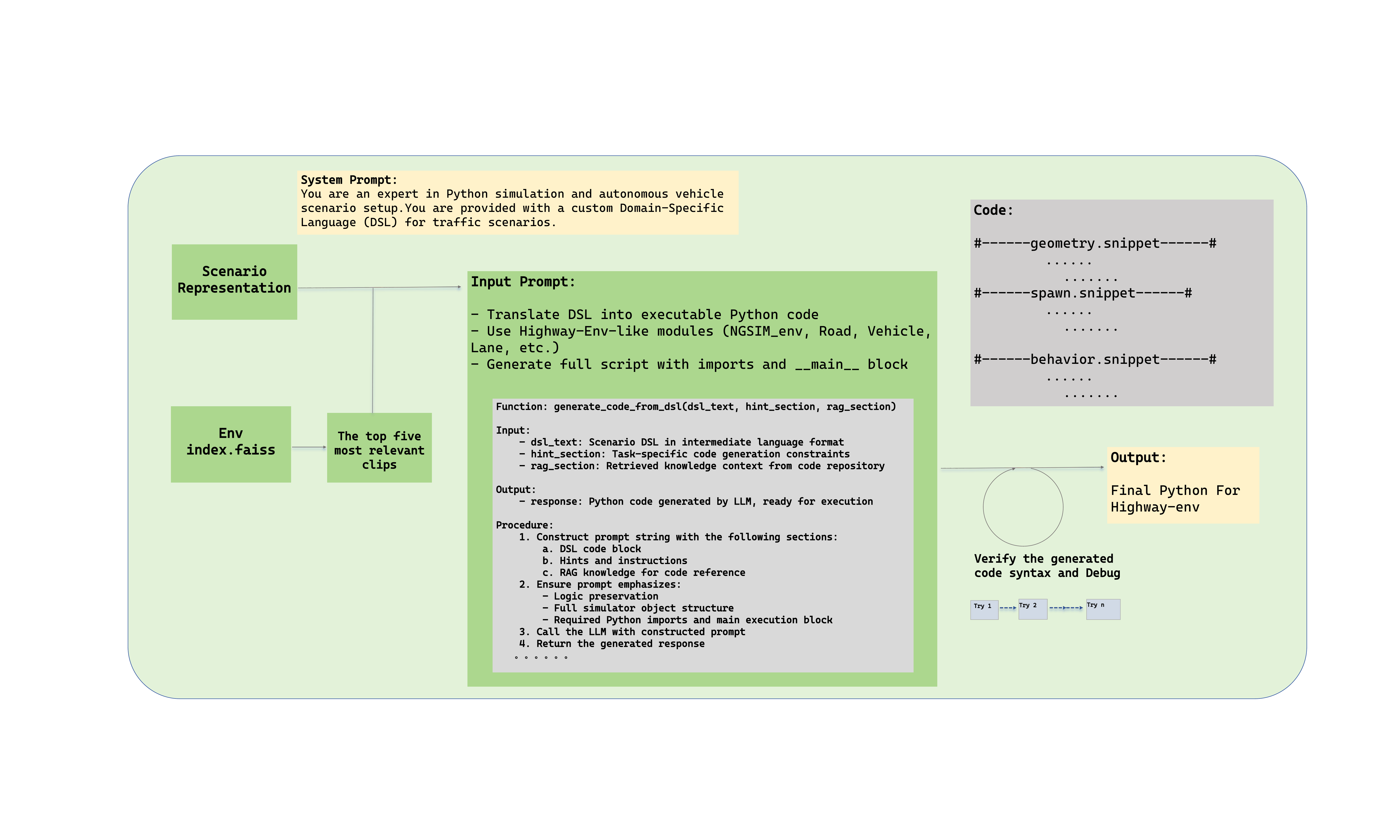}
  \caption{Workflow of DSL-to-Python code generation: retrieval of relevant code fragments, prompt assembly, LLM-based code generation, and iterative syntax validation.}
  \label{fig:example3}
\end{figure*}

With the repository prepared, we proceed to detail the process of translating DSL representations into executable simulation codes.  
The overall workflow for DSL-to-Python code generation, including retrieval, prompt construction, and iterative debugging, is illustrated in Figure~\ref{fig:example3}.

The code generation pipeline employs a RAG method, combined with automatic syntax validation and iterative correction mechanisms, ensuring that the final output adheres to both semantic and syntactic constraints.
The overall pipeline is illustrated in Algorithm~\ref{alg:dsl2code_main}.

\begin{algorithm}[th]
\caption{DSL to Python Code Generation with RAG and Iterative Debugging}
\label{alg:dsl2code_main}
\KwIn{DSL string $d$}
\KwOut{Executable Python code file}

\SetKwFunction{FPrompt}{build\_conversion\_prompt}
\SetKwFunction{FLLM}{call\_llm\_for\_code}
\SetKwFunction{FValidate}{validate\_python\_code}
\SetKwFunction{FSave}{save\_code\_to\_file}

\BlankLine
Load FAISS index of Python code corpus: $index \leftarrow$ \texttt{load\_highway\_code\_index()}\\
$context \leftarrow$ Top-$k$ code snippets from $index$ similar to $d$\\
Initialize $fix\_hint \leftarrow \texttt{None}$, $attempt \leftarrow 1$

\BlankLine
\While{$attempt \le 3$}{
    $prompt \leftarrow$ \FPrompt{$d$, $context$, $fix\_hint$}\\
    $raw\_code \leftarrow$ \FLLM{$prompt$} \\
    Clean code with \texttt{strip\_code\_block} and \texttt{clean\_start} \\

    \If{\FValidate{code} is valid}{
        \FSave{code, path} \\
        \Return code
    }
    \Else{
        $fix\_hint \leftarrow$ extract error msg\\
        $attempt \leftarrow attempt + 1$
    }
}
\Return ``Generation failed''
\end{algorithm}

\paragraph{Prompt Construction.}
A structured prompt is assembled by embedding three main elements: (i) the target DSL to be converted, (ii) context snippets retrieved from the code repository, and (iii) optional syntax error correction hints derived from previous generation attempts.  
The prompt enforces the following generation constraints:
\begin{itemize}
    \item Incorporate all necessary import statements and class references.
    \item Preserve logical and behavioral dependencies among scenario components.
    \item Format the script according to Python 3.8+ syntax conventions.
    \item Organize the code into modular functions, including a \texttt{main()} entry point for execution.
\end{itemize}
The prompt structure utilized for DSL-to-Python code generation is detailed in Table~\ref{tab:dsl2python_prompt_response} and the detailed construction logic is described in Algorithm~\ref{alg:build_conversion_prompt} 

\begin{table*}[t]
\centering
\caption{Example of prompt design for DSL-to-Python code generation with RAG context and iterative debugging.}
\label{tab:dsl2python_prompt_response}
\begin{tcolorbox}[colback=white!5!white, colframe=black!75!black, title=Prompt:]

\textbf{System Prompt:}  
You are an expert in Python simulation and autonomous vehicle scenario setup. Your task is to translate the given Domain-Specific Language (DSL) representation of a traffic scenario into a complete and executable Python script, compatible with a customized extension of the \texttt{highway-env} simulator.

\textbf{Input Prompt:}
\begin{enumerate}
    \item \textbf{Scenario Context:}  
    Optionally append top-$k$ similar code snippets retrieved from the FAISS-based repository as additional context to guide generation.

    \item \textbf{Instructions for Code Generation:}
    \begin{itemize}
        \item Preserve all logic encoded in the DSL behaviors without truncation.
        \item Convert regions and vehicle setups into Python objects using the simulator's modules.
        \item Define all scenario parameters, vehicle spawn positions, and behaviors appropriately.
        \item Format the output cleanly with proper imports and consistent spacing.
        \item Include a \texttt{if \_\_name\_\_ == '\_\_main\_\_'} block to serve as an execution entry point.
        \item Ensure necessary imports at the beginning of the script, including modules such as:
\begin{verbatim}
from NGSIM_env import utils
from NGSIM_env.envs.common.abstract import AbstractEnv
from NGSIM_env.road.road import Road, RoadNetwork
from NGSIM_env.vehicle.behavior import IDMVehicle
...
\end{verbatim}
        \item Ensure the generated code compiles under Python 3.8+ without syntax errors.
    \end{itemize}

    \item \textbf{Syntax Error Correction (Optional):}  
    If previous generation attempts failed due to syntax errors, prepend the extracted error message as a corrective hint to guide the next generation.

    \item \textbf{DSL Embedding:}
    \begin{itemize}
        \item Include the full DSL content within the prompt, properly formatted as a code block.
    \end{itemize}
\end{enumerate}

\textbf{Prompt Sections Structure:}

[Instructions for DSL Conversion]

[RAG Context: Retrieved Similar Code Fragments]

[Syntax Fix Hint]

[DSL Content Block]

\end{tcolorbox}

\vspace{0.5em}

\begin{tcolorbox}[colback=white!5!white, colframe=black!75!black, title=Response:]

\textbf{Generated Output:}
\begin{verbatim}
#------ Python Scenario Script ------#
- Import required modules
- Initialize road network and map
- Create ego vehicle and surrounding vehicles
- Configure vehicle behaviors and interaction logic
- Assemble scenario components
- Launch simulation with entry point
\end{verbatim}
\end{tcolorbox}
\end{table*} %

\begin{algorithm}[th]
\caption{build\_conversion\_prompt: Prompt Construction with Context and Hints}
\label{alg:build_conversion_prompt}
\KwIn{DSL $d$, context snippets $C$, syntax fix hint $h$}
\KwOut{Prompt string $P$}

\BlankLine
Initialize prompt with DSL conversion instruction block\\
Append context section using top-$k$ code snippets $C$\\
\If{$h \neq$ None}{
    Append error fix hint section to prompt
}
Append detailed instructions: imports, spacing, structure, `main` block\\
Append DSL as code block formatted string\\
\Return formatted prompt $P$
\end{algorithm}

\paragraph{Syntax Validation and Debugging.}
After each generation attempt, the output Python code undergoes syntax validation using the Python \texttt{compile()} function. If the code passes without errors, it is saved as a complete script for downstream simulation. If syntax errors are detected, the system extracts the error message and incorporates it as a correction hint into the next prompt iteration.  
This debugging cycle is repeated for a limited number of attempts (typically three) to maximize success chances while avoiding infinite correction loops.  
The syntax checking and debugging procedure is formalized in Algorithm~\ref{alg:validate_python_code}.

\begin{algorithm}[th]
\caption{validate\_python\_code: Syntax Checking of LLM Output}
\label{alg:validate_python_code}
\KwIn{Code string $c$}
\KwOut{Validity flag, error message}

\BlankLine
Try compiling $c$ using Python \texttt{compile()} \\
\If{compile success}{
    \Return True, ``''
}
\Else{
    Extract error line and message\\
    \Return False, error string
}
\end{algorithm}

\paragraph{Final Code Output.}
Upon successful validation, the generated Python script includes the following elements:
\begin{itemize}
    \item Map structure initialization and lane network definitions.
    \item Ego vehicle and surrounding adversarial vehicle instantiation.
    \item Behavior controller configurations for dynamic scenario execution.
    \item Scenario setup and simulation startup entry points.
\end{itemize}

The final executable code is compatible with the customized \texttt{highway-env} simulation environment and can be directly rendered and tested using the simulator’s visualization and control modules.

\section{PPO-based Agent Training Method}
\label{appendix:ppo_training}

To enable realistic, dynamically challenging driving scenarios aligned with user-defined descriptions, we implement a PPO-based reinforcement learning framework that alternates between training and freezing adversarial agents' weights.  
Specifically, adversarial behaviors are initialized according to scenario representations derived from structured DSL modules, ensuring consistency with the original user intent.

The training process is organized following an "open-closed" principle:  
During the \textbf{open phase}, adversarial vehicles are actively trained using PPO to maximize disruption towards ego vehicles while maintaining realistic driving behaviors.  
After adversarial agents reach satisfactory performance, their policy parameters are \textbf{closed} — frozen — and ego vehicles are subsequently trained against these fixed adversaries to enhance robustness and safe maneuvering.

In the open phase, adversarial agents are encouraged to interfere with ego trajectories via behaviors consistent with their DSL-defined templates, such as aggressive lane cutting, abrupt braking, or forced lateral conflicts.  
The adversarial reward function incentivizes successful disruptions (e.g., induced collisions, forced emergency maneuvers) while penalizing physically implausible behaviors.  
To ensure sustained exploration and avoid premature convergence to local optima, we incorporate a reset mechanism, where agent parameters and environmental seeds are periodically reinitialized every 5000 environment steps.  
This dynamic resetting encourages continual discovery of new adversarial strategies and scenario variations.

Once adversarial agents demonstrate stable disruptive capabilities, their policy networks are frozen, initiating the close phase.  
In this phase, ego vehicles are trained against the now-fixed adversarial behaviors, focusing on maximizing survival rates, maintaining efficient trajectories, and adhering to intended driving goals.  
The ego vehicle reward structure rewards collision avoidance, smooth control, and goal-directed navigation, while penalizing unsafe behaviors and unnecessary conservatism.

Both training phases employ standard PPO with clipped surrogate loss, entropy regularization, and generalized advantage estimation (GAE).  
The policy and value networks are implemented as two-layer MLPs with 256 hidden units per layer and ReLU activations, optimized using Adam with a learning rate of $3 \times 10^{-4}$.  
The rollout length is set to 2048 steps, and mini-batches of size 256 are used for updates.

Throughout training, interaction trajectories are logged and selectively integrated into the adversarial scenario corpus.  
Successful evasions, near-miss events, and collision cases enrich the dataset, progressively augmenting the diversity and difficulty of the available evaluation scenarios.

This "open-then-close" PPO training paradigm, tightly coupled with user-defined DSL scene specifications and enhanced by continual resetting, enables the generation of high-fidelity, executable, and behaviorally diverse driving scenarios critical for robust autonomous agent evaluation.

\section{Trajectory Optimization via Bézier Curves}
\label{appendix:bezier}

To refine the trajectories produced in the simulation environment and ensure both safety and smoothness, we introduce a post-processing module based on Bézier curve optimization. This module transforms discrete vehicle trajectories into continuous, differentiable paths that more closely approximate realistic driving behavior while avoiding collisions and minimizing abrupt control actions.

Bézier curves are parametric curves widely used in computer graphics and robotics due to their controllable shape and inherent smoothness. A Bézier curve of degree $n$ is defined by a set of $n+1$ control points, where each point contributes to the curve's shape through a weighted sum of Bernstein polynomials. Given a temporal parameter $t \in [0,1]$ and a sequence of control points $\{P_0, P_1, \dots, P_n\}$, the Bézier curve $\mathbf{B}(t)$ is expressed as:
\[
\mathbf{B}(t) = \sum_{i=0}^{n} \binom{n}{i} (1 - t)^{n-i} t^i P_i
\]

In our framework, the initial set of control points is extracted from the raw trajectory produced by the ego vehicle within the simulation. These are selected either uniformly across the temporal sequence or centered around critical trajectory segments such as intersections or lane changes. A temporal sampling vector is generated to align the curve with the original time sequence.

We formulate the optimization of control points as a convex objective that minimizes the deviation between the optimized Bézier trajectory and the original trajectory. Let $\{\mathbf{x}_t, \mathbf{y}_t\}$ denote the position of the ego vehicle at time $t$ in the original trajectory and let $\{\hat{\mathbf{x}}_t, \hat{\mathbf{y}}_t\}$ be the Bézier-evaluated positions. The cost function is defined as the sum of squared errors:
\[
\mathcal{L} = \sum_t \left( (\mathbf{x}_t - \hat{\mathbf{x}}_t)^2 + (\mathbf{y}_t - \hat{\mathbf{y}}_t)^2 \right)
\]
We use the L-BFGS-B optimization algorithm to minimize this loss, adjusting the flattened vector of control point coordinates under optional boundary and smoothness constraints. This process significantly improves trajectory continuity, reduces high-curvature segments and enables smoother vehicle control actions in subsequent simulations. Moreover, the optimized Bézier trajectories retain alignment with the semantic intent of the original scene, ensuring consistency with both behavioral design and downstream safety evaluation.

\section{Experiment Setup}
\label{appendix:experiment-setup}

\subsection{Simulation Environment Configuration}

Our simulation environment is built upon a customized version of \texttt{highway-env}, extended to support a full pipeline of natural language-driven scenario generation, DSL compilation and closed-loop agent interaction. This environment serves as the execution backbone for testing and refining autonomous driving policies in various traffic situations that are critical to safety.

The pipeline begins with a structured corpus of annotated driving scenarios, which is indexed using FAISS to support semantic retrieval. Given a user-provided natural language description, our system employs a RAG approach: it retrieves relevant repository entries and uses chain-of-thought prompting with a LLM to produce a structured DSL representation. This stage is implemented using a self-consistent voting mechanism to ensure output stability and semantic fidelity.

The resulting DSL, which includes geometric layout, vehicle spawn locations and behavioral definitions, is translated into executable code via a dedicated compiler module. This module leverages a secondary FAISS index built over the highway-env codebase, allowing code generation to be informed by retrieval-based context examples. Syntax validation and multi-round debugging are performed iteratively until valid, executable simulation scripts are produced.

The final simulation scripts instantiate road networks and agent behaviors within the modified highway-env framework. To support learning-based evaluation, we integrate a trained autonomous agent capable of generalizing across the generated adversarial scenarios. Furthermore, a Bézier curve convex optimization module refines the ego vehicle's trajectories to ensure safety, continuity and control smoothness.

\subsection{Agent Setup and Control Policy}

To evaluate the robustness and safety of autonomous behaviors under adversarial scenarios, we train a RL agent using the PPO algorithm. The agent takes as input a high-level representation of the driving scene, which includes lane topology, positions of surrounding objects and behavioral indicators. Based on these observations, the agent outputs discrete control actions such as lane changes, acceleration and emergency braking. The reward function is carefully designed to penalize collisions and abrupt maneuvers, while promoting smooth and goal-directed trajectories.

To ensure that the ego vehicle maintains realistic and safe motion even under adversarial conditions, we integrate a post-processing module that applies Bézier curve-based convex optimization. This module takes the raw action sequence output by the PPO policy and transforms it into a continuous, differentiable trajectory. The optimization minimizes curvature spikes and maintains temporal consistency with the original simulation time steps.

PPO agent is trained by using a curriculum of DSL-generated scenarios, ranging from benign to highly adversarial interactions. This training approach enables the agent to generalize across a wide variety of scene configurations, improving its ability to react safely to unexpected hazards. 
The final agent is evaluated though using collision rates across the entire scenario corpus.

\subsection{Scenario Generation Process}

To facilitate natural language-based scenario specification, we design a two-stage generation pipeline that converts free-form descriptions into executable simulation inputs. This pipeline consists of: (1) a RAG mechanism for generating structured DSL scenes from text and (2) a dedicated compiler that transforms DSL into Python code compatible with our simulation environment.

The RAG stage is powered by a multi-layered scenario corpus constructed from real-world trajectory datasets. Each entry contains a natural language description, structured DSL snippets (geometry, spawn and behavior) and metadata such as scenario category and vehicle interaction types. These entries are embedded using a Sentence-BERT model and stored in a FAISS index, enabling fast semantic retrieval. Upon receiving a user query, the system retrieves the top-k most relevant repository entries to construct an informed prompt for the LLM.

We use CoT prompting combined with few-shot examples to guide the LLM in producing a well-structured DSL output. To improve consistency and mitigate hallucination, we implement a self-consistency voting strategy: the same query is sampled multiple times and the outputs are ranked via embedding similarity. The most representative DSL is selected for downstream compilation.
The second stage involves translating the finalized DSL into executable Python code using our internal module. This module utilizes a secondary FAISS index built over a curated set of highway-env code examples. The conversion process is guided by retrieval-based prompt injection and follows strict syntax validation and correction loops until a valid scenario script is produced.

\subsection{Scoring Criteria for Comparison and Ablation Studies}

To systematically evaluate the quality of the generated DSL representations and the corresponding code outputs, we design a structured rubric consisting of six criteria. Each criterion is individually scored on a scale from 0 (very poor) to 5 (excellent), and weighted according to its importance for realistic autonomous driving simulation. The evaluation criteria are summarized in Table~\ref{tab:dsl_metrics}.

\begin{table*}[t]
\centering
\caption{Evaluation Criteria for DSL Scenario Quality}
\label{tab:dsl_metrics}
\begin{tabular}{p{3.8cm}p{8.8cm}p{1.5cm}}
\toprule
\textbf{Criterion} & \textbf{Description} & \textbf{Weight} \\
\midrule
Semantic Fidelity & Measure the alignment between the DSL and the input scenario description, including scene type, number of vehicles, positions, behaviors and interaction logic. & 25\% \\
Executable Validity & Check whether the code is grammatically correct and executable under our simulation framework. & 20\% \\
Structural Completeness & Verify the presence and integrity of the three modules—\texttt{geometry}, \texttt{spawn} and \texttt{behavior}. & 15\% \\
Modularity and Maintainability & Evaluate code reuse, naming clarity and absence of hardcoded logic, indicating engineering quality and extensibility. & 15\% \\
Behavioral Modeling Richness & Assess the vehicle behaviors modeled in detail, rather than using vague or default actions. & 20\% \\
Voting Centrality Score & Measure the proximity of the final DSL to the centroid of multiple sampled candidates, reflecting generation stability. & 5\% \\
\bottomrule
\end{tabular}
\end{table*}

Based on the assigned weights, a weighted total score is computed and normalized to a 100-point scale.  
This DSL scenario quality score is utilized throughout our experiments to enable both cross-model comparisons of different LLM backbones and rigorous ablation-based validation of critical design components, including few-shot prompting, chain-of-thought guidance, semantic alignment strategies, and self-consistency voting mechanisms.

\section{Supplementary Notes on Experiment Results}
\label{appendix:experiment_result_details}

\paragraph{DSL Generation Evaluation Procedure.}
Each model (DeepSeek-V3, DeepSeek-R1, Qwen2.5-Plus) is evaluated across six ego behaviors: \textit{Brake/Following}, \textit{Lane Change}, \textit{Go Straight}, \textit{Turn Left}, \textit{Turn Right}, and \textit{Turn Round}.  
For each behavior:
\begin{itemize}
    \item Five independent runs are conducted, each generating a distinct DSL output using the full generation pipeline.
    \item Each DSL is evaluated using the structured rubric defined in Table~\ref{tab:dsl_metrics}.
    \item Scoring is performed automatically via GPT-3.5-turbo or GPT-4 (denoted as GPT-o3), following a fixed structured evaluation prompt.
    \item The five scores are averaged to obtain the final score for that behavior and model.
\end{itemize}

After evaluating all six behaviors, the overall model performance is summarized by computing the macro-average across all behavior types, as shown in Table~\ref{tab:llm_dsl_comparison}.
No model-specific prompt adjustments are introduced. Random seeds are fixed during sampling to ensure reproducibility.
\begin{table}[ht]
\centering
\caption{Comparison of DSL scenario generation scores across different LLMs under various ego behaviors. Higher scores indicate better semantic fidelity and structural completeness.}
\label{tab:llm_dsl_comparison}
\begin{tabular}{lccc}
\toprule
\textbf{Ego Behavior} & \textbf{DeepSeek-V3} & \textbf{Qwen2.5-Plus} & \textbf{DeepSeek-R1} \\
\midrule
Brake / Following  & 80 & 65 & 78 \\
Lane Change        & 75 & 58 & 75 \\
Go Straight        & 81 & 72 & 75 \\
Turn Left          & 81 & 58 & 79 \\
Turn Right         & 88 & 66 & 68 \\
Turn Round         & 76 & 61 & 68 \\
\midrule
\textbf{Average}   & \textbf{80.17} & \textbf{63.33} & \textbf{73.83} \\
\bottomrule
\end{tabular}
\end{table}

\paragraph{Ablation Study Settings.}
In the ablation experiments, DeepSeek-V3 is used as the backbone LLM.  
Two core generation components are selectively disabled:
\begin{itemize}
    \item \textbf{Self-Consistency Voting:} Only a single LLM output is used without voting.
    \item \textbf{Semantic Alignment Checking:} Semantic validation between input description and DSL output is omitted.
\end{itemize}

The same set of scenarios, prompts, retrieval corpus, and scoring procedures is maintained to ensure a fair comparison with the full pipeline results.  
The reported ablation results in Table~\ref{tab:ablation} thus isolate the effects of each module while controlling other variables.

\paragraph{Collision and Timeout Evaluation Protocol.}
The reinforcement learning experiments evaluate the ego vehicle's robustness against adversarial agents under generated DSL scenarios.  
Training and evaluation proceed as follows:
\begin{itemize}
    \item PPO agents are trained using centralized multi-process rollouts over adversarial scenarios.
    \item Each adversarial scenario type (braking/following, lane changing, going straight, turning) includes at least 50 samples.
    \item Ego behavior is evaluated both with fixed scripted policies (baseline) and with reinforcement-learned avoidance policies.
\end{itemize}

Collision rates reflect the proportion of simulation runs ending in ego vehicle failure due to adversarial actions.  
Timeout counts reflect survivals until the simulation horizon is reached without collision.  
Statistics are aggregated and averaged across scenarios for each ego behavior type, as reported in Table~\ref{tab:collision-stats}.

\paragraph{In-depth Analysis of Experimental Results.}

\begin{itemize}
    \item \textbf{Cross-model DSL Generation Performance.}  
DeepSeek-V3 consistently outperforms DeepSeek-R1 and Qwen2.5-Plus across all ego behaviors.  
Its advantage is particularly pronounced in structurally complex scenarios such as \textit{Turn Left} and \textit{Turn Right}, suggesting a stronger capacity for modeling fine-grained semantic relationships and interaction logic.  
The performance gap between DeepSeek-V3 and Qwen2.5-Plus (approximately 17 points on average) highlights the critical role of LLM backbone quality in structured autonomous driving scenario generation.

    \item \textbf{Ablation Study Findings.}  
The full pipeline, which incorporates few-shot prompting, CoT reasoning, semantic alignment, and self-consistency voting, yields the highest DSL generation quality.  
Removing the voting module causes moderate degradation, primarily affecting output stability rather than semantic accuracy.  
In contrast, removing semantic alignment causes sharper declines, particularly in scenarios requiring intricate multi-agent coordination, confirming its importance for ensuring DSL-output fidelity to input descriptions.

    \item \textbf{Collision and Timeout Outcomes.}  
Reinforcement learning significantly reduces collision rates across all behavior categories.  
In particular, \textit{Brake} and \textit{Turn} scenarios show the most substantial improvements, reflecting the effectiveness of adaptive avoidance strategies against dynamic adversarial threats.  
Timeout rates remain relatively stable, indicating that the learned policies maintain operational efficiency while enhancing safety.
\end{itemize}

Overall, these findings validate the effectiveness of retrieval-augmented, prompt-guided DSL generation pipelines in producing both semantically accurate and simulation-executable autonomous driving scenarios.

\end{document}